\pdfoutput=1
\documentclass[11pt]{article}

\usepackage{ACL2023}

\usepackage{times}
\usepackage{latexsym}

\usepackage[T1]{fontenc}
\usepackage[utf8]{inputenc}
\usepackage{booktabs}
\usepackage{graphicx}  % for including images
\usepackage{float}     % for positioning images
\usepackage{hyperref}
\usepackage{footmisc}
\usepackage{array}
\usepackage{enumitem}
\usepackage{algorithm}
\usepackage{algorithmic}
\usepackage{amsmath}
\usepackage{graphicx}
\usepackage{todonotes}
\usepackage{color,soul}
\usepackage{microtype}

\usepackage{inconsolata}

\title{Recon, Answer, Verify: Agents in Search of Truth}

\author{
  Satyam Shukla \hspace{1em} Himanshu Dutta \hspace{1em} Pushpak Bhattacharyya \\
  Indian Institute of Technology Bombay \\
  \texttt{\{satyamshukla, himanshud, pb\}@cse.iitb.ac.in}
}

\begin{document}
\maketitle
\begin{abstract}
Automated fact-checking with large language models (LLMs) offers a scalable alternative to manual verification. Evaluating fact-checking is challenging as existing benchmark datasets often include post-claim analysis and annotator cues, which are absent in real-world scenarios where claims are fact-checked immediately after being made. This limits the realism of current evaluations. We present \textbf{Politi-Fact-Only (PFO)}, a 5-class benchmark dataset of 2,982 political claims from politifact.com\footnote{\url{https://www.politifact.com/}}, where all post-claim analysis and annotator cues have been removed manually. This ensures that models are evaluated using only the information that would have been available prior to the claim’s verification. Evaluating LLMs on PFO, we see an average performance drop of \textbf{22\%} in terms of macro-f1 compared to PFO's unfiltered version. Based on the identified challenges of the existing LLM-based fact-checking system, we propose \textbf{RAV (Recon-Answer-Verify)}, an agentic framework with three agents: question generator, answer generator, and label generator. Our pipeline iteratively generates and answers sub-questions to verify different aspects of the claim before finally generating the label. RAV generalizes across domains and label granularities, and it outperforms state-of-the-art approaches on well-known baselines RAWFC \textit{(fact-checking, 3-class)} by \textbf{25.28\%}, and on HOVER \textit{(encyclopedia, 2-class)} by \textbf{1.54\%} on 2-hop, \textbf{4.94\%} on 3-hop, and \textbf{1.78\%} on 4-hop, sub-categories respectively. RAV shows the least performance drop compared to baselines of \textbf{16.3\%} in macro-f1 when we compare PFO with its unfiltered version.
\end{abstract}

%claim are in the abstract
%qualitative analysis
%appendix should be compact
%
\section{Introduction}

\begin{figure}[t]
    \centering
    \includegraphics[width=1\linewidth, trim=0.5cm 1cm 0.5cm 0.5cm, clip]{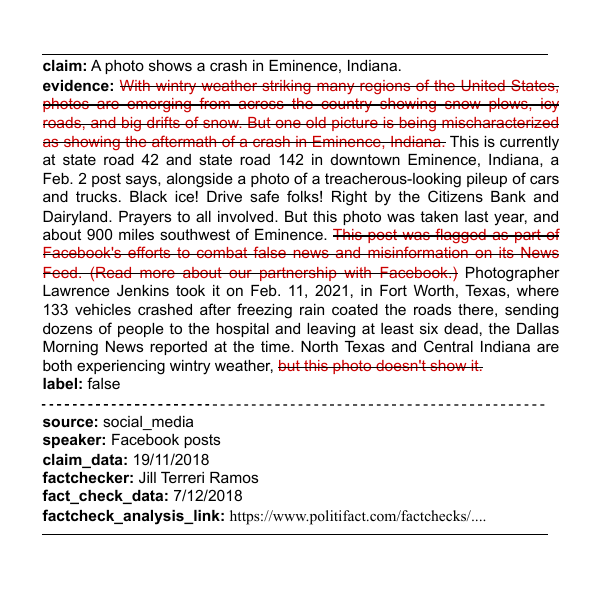}
    \caption{An instance from the PFO dataset where sentences with strike-through represent information added only after the claim was verified (post-publication analysis or annotator cues). These lines were manually deleted to ensure the evidence contained only factual details related to the claim. We also show the metadata available for an instance in PFO.}
    \label{fig:instance_data}
\end{figure}

% Several studies have looked into automated fact-checking, including work by \citet{wang-2017-liar}, \citet{augenstein-etal-2019-multifc}, and \citet{gupta-srikumar-2021-x}, who have created useful datasets. These datasets include real-world claims but mostly come from fact-checking websites. Articles on these sites often include analysis done after the claim was made \textbf{(leakage)}, where annotators add their judgments using facts. This setup does not fully match how fact-checking happens in real life. Rather than using articles published on fact-checking websites, some researchers, like \citet{yang2022cofced} and \citet{khan-etal-2022-watclaimcheck}, use source articles that were published before the claim. This helps make the process more realistic. Still, even though claims are matched with relevant parts of these documents, the content found might not always be enough to verify the claim.

Several studies have explored automated fact-checking, including work by \citet{wang-2017-liar}, \citet{augenstein-etal-2019-multifc}, and \citet{gupta-srikumar-2021-x}, who created widely-used benchmark datasets. While these datasets contain real-world claims, they are primarily derived from fact-checking websites, where each claim is often accompanied by detailed analysis and verdicts written after the claim was made. This post-claim analysis frequently includes annotator judgments or interpretive statements that explicitly indicate the truthfulness of the claim, what we refer to as \textbf{leakage}. Leakage, also called post-claim analysis or annotator cues, provides information that would not be available in a real-time fact-checking scenario and can inadvertently guide model predictions. To address this, some researchers, such as \citet{yang2022cofced} and \citet{khan-etal-2022-watclaimcheck}, have instead used contemporaneous source articles documents published before the claim to simulate more realistic settings. However, even when using pre-claim documents, the evidence may still be insufficient or too loosely related to support or refute the claim effectively, limiting the reliability of model evaluation.

% However, even when claims are linked to such prior documents, they may still be paired with text that indirectly reflects post-claim interpretation or includes insufficient context for verification.

To address this problem, \textit{we propose} a benchmark dataset in the political domain, \textbf{Politi-Fact-Only (PFO)} (see Section~\ref{sec:dataset}), a subset of \citet{misra2022politifact}. In this dataset, we manually remove post-claim analysis and annotator cues provided by fact-checkers. For example, phrases such as \textit{``But one old picture is being mischaracterized as showing the aftermath of a crash in Eminence, Indiana.''} and \textit{``but this photo doesn't show it''} (see Figure~\ref{fig:instance_data}) directly signal that the claim is false. Such cues are unavailable in real-world scenarios, where evidence typically consists of content published prior to the claim itself. In PFO, we retain only pre-claim content, ensuring that fact-checking models are evaluated under conditions that more closely resemble real-time verification.

The rapid spread of misinformation on digital platforms, especially social media, has intensified the need for domain-agnostic fact-checking systems \cite{vosoughi2018spread}. Manual verification is slow and resource-intensive \cite{graves2018understanding}, and less than half of published articles undergo any fact-checking at all \cite{lewis2008quality}. This creates a pressing demand for automated approaches that can scale across domains and support fine-grained labelling. One such effort is ProgramFC \cite{pan2023factchecking}, which uses program-guided reasoning, encoding verification steps as intermediate executable programs. Labels are predicted based on the logical truth of Boolean expressions: \textit{true} indicates support, and \textit{false} indicates refutes. However, this approach is restricted to binary classification, limiting its applicability to more nuanced multi-class fact-checking scenarios. \citet{zhang2023llmbased} worked on decomposing a claim into sub-components and generating questions to verify each of the components. Their study is limited to the political domain. They used multiple in-context examples from the political domain, and their question generation process is based on only those in-context examples. As no task description is provided, it becomes limited to the political domain. 

To address this challenge, \textit{we propose} Recon-Answer-Verify \textbf{(RAV)}, a generalized agentic framework. Our approach starts with a Question Generator agent (QG\textsubscript{agent}), which produces a sub-question and brief reasoning based on the claim and earlier sub-questions. The reasoning links the claim with previous questions and answers to identify parts that are still unverified. Then, a related sub-question is generated. This makes the process interpretable and applicable across domains. The Answer Generator agent (AG\textsubscript{agent}) answers each question using the evidence provided with the claim. The process is iterative: at each step, the QG\textsubscript{agent} checks whether all aspects of the claim are covered or more questions are needed. Once the claim is fully explored, the Label Generator agent (LG\textsubscript{agent}) generates reasoning that connects the questions, answers, and claim, and then predicts the final label. Reasoning is generated during both question and label steps, making fact-checking more transparent.
\\

\noindent Our contributions are:
\begin{enumerate}[leftmargin=*]
    \item \textbf{Politi-Fact-Only (PFO)} (Section~\ref{sec:dataset}): We introduce a benchmark dataset of \textbf{2,982} instances evenly distributed across five classes: true, mostly-true, half-true, mostly-false, and false. We manually remove all post-claim analyses and annotator commentary, keeping only factual evidence (see Fig. \ref{fig:instance_data}) to ensure unbiased input and better grounding in reality. When evaluated on PFO as opposed to its unfiltered version. LLM show a performance drop of \textbf{22\%} on average (see Table \ref{table:model_comparison_macro}). Content length reduction confirms that \textbf{17.79\%} of the original data consisted of commentary or verdict cues (see Table~\ref{table:token_content_drop}).

    \item \textbf{Recon-Answer-Verify (RAV)} (Section~\ref{sec:methodology}): An agentic pipeline that breaks fact verification into a question-answering process. RAV uses domain-agnostic prompts, allowing it to generalize across domains and label granularities. It outperforms state-of-the-art baselines on RAWFC (3-class, multi-domain) by \textbf{25.28\%}, and on HoVer (2-class, encyclopedia) by \textbf{1.54\%} (2-hop), \textbf{4.94\%} (3-hop), and \textbf{1.78\%} (4-hop). RAV shows the least performance drop of \textbf{16.3\%} in macro-f1, when we evaluate it on PFO as opposed to its unfiltered version (Table~\ref{tab:main_comparison_variant4}).

    \item \textbf{Comprehensive evaluation of the RAV pipeline} (Section~\ref{subsec:varients}): We systematically study the effect of different question-generation strategies and question types: Verification (True/False answerable) and inquiry (requires explanation) question on the fact-checking task (see Table~\ref{tab:ablation_models_first}). We conduct experiments with LLMs ranging from 7B to 70B parameters, and analyse how performance over the fact-checking task is affected by the scale of LLMs. We show that the final RAV pipeline, designed to reason \emph{iteratively} and generate both \emph{verification and inquiry questions}, achieves better performance compared to other variants (see Table~\ref{tab:main_comparison_variant4}).
\end{enumerate}

\section{Problem Statement}
% format like input and output
The problem of veracity detection can be stated as: Given a claim \texttt{C} and corresponding evidence \texttt{E}, finding the corresponding veracity label \( y \in G \), where \( G \) can be either two-class, three-class or five-class sets of veracity labels. The prediction is \(y^* = \arg\max_{y \in G} P(y \mid C, E)\).

\section{Related Work}
\label{sec:relatedwork}

% Fact-checking datasets vary significantly in the quality and nature of their evidence, especially with respect to leakage such as annotator commentary and verdict cues. Datasets like LIAR \cite{wang-2017-liar} and the dataset by \citet{rashkin-etal-2017-truth} lack supporting textual evidence, which limits their applicability in evidence-based verification. In contrast, FEVER \cite{thorne-etal-2018-fever} and HOVER \cite{jiang-etal-2020-hover} rely exclusively on Wikipedia passages as evidence, resulting in datasets that are largely free from leakage due to the controlled and curated nature of their sources. 

% Datasets such as MultiFC \cite{augenstein-etal-2019-multifc} and X-Fact \cite{gupta-srikumar-2021-x} incorporate evidence from broader, less curated sources, making them more prone to leakage, including annotator comments and verdict-related signals. Similarly, LIAR-PLUS \cite{alhindi-etal-2018-evidence} contains evidence extracted from source articles, often with incomplete context that may contribute to leakage or insufficient evidence. Fact-checking datasets derived from fact-checking websites, including L++ \cite{10.1162/tacl_a_00601} and RU22fact \cite{zeng-etal-2024-ru22fact}, also tend to contain leaked commentary and verdict cues embedded within their evidence texts.

Fact-checking datasets differ widely in the quality and reliability of their evidence, particularly regarding leakage such as annotator commentary or verdict cues. Early datasets like LIAR \cite{wang-2017-liar} and \citet{rashkin-etal-2017-truth} lack explicit supporting evidence, limiting their use in evidence-based verification. Others, such as FEVER \cite{thorne-etal-2018-fever} and HOVER \cite{jiang-etal-2020-hover}, avoid leakage by using only curated Wikipedia content. In contrast, datasets like MultiFC \cite{augenstein-etal-2019-multifc}, X-Fact \cite{gupta-srikumar-2021-x}, and LIAR-PLUS \cite{alhindi-etal-2018-evidence} draw from broader or less curated sources, often introducing leakage through annotator comments, verdict hints, or incomplete context. Similarly, datasets sourced from fact-checking websites, including L++ \cite{10.1162/tacl_a_00601} and RU22fact \cite{zeng-etal-2024-ru22fact}, frequently embed verdict-related cues within their evidence texts, further limiting their realism for modelling real-world fact-checking.
Apart from dataset construction, recent work has focused on improving fact verification through advanced reasoning and prompting strategies. \citet{yang2022cofced} introduced CofCED, a supervised framework that first retrieves relevant evidence from a candidate pool and then applies evidence distillation and refinement to enhance claim verification accuracy. Addressing real-world scenarios, \citet{khan-etal-2022-watclaimcheck} focused on sourcing content that predates the claim publication to improve evidence relevance. \citet{cohen2023lm} proposed a cross-examination framework where two language models interact to mitigate hallucination.

\section{Politi-fact-only: A Fact Only Benchmark Dataset}
\label{sec:dataset}
\textit{Politi-Fact-Only} is a curated dataset that ensures each claim's evidence contains only factual information. It is created from a subset of the PolitiFact dataset \cite{misra2022politifact}. On this contains around 21k instances, where we applied a filtration process to 3000 randomly selected instances, with nearly 600 instances from each class. After the filtration process, the final dataset consists of 2,982 instances (explained in Section~\ref{sec:annotation}). Each instance in \textit{PFO} dataset contains nine attributes: \textit{label}, \textit{claim}, \textit{evidence}, \textit{speaker}, \textit{factcheck\_analysis\_link}, \textit{factcheck\_date}, \textit{fact\_checker}, \textit{claim\_date}, and \textit{claim\_source}. Additionally, the \textit{false} and \textit{pants-fire} labels were merged into a single \textit{false} label, as both categories represent completely false information. Examples from each class can be found in the Appendix~\ref{appex:pfo_examples}. We retain all attributes of the PolitiFact dataset, and we add one more attribute: \emph{evidence}. We further split the PFO dataset into three parts: the training set, the test set, and the validation set, containing 2482, 300, and 200 instances, respectively.
\begin{table}[h]
    \centering
    \small
    \begin{tabular}{l|c|cc|c}
        \textbf{Label} & \textbf{Count} & \multicolumn{2}{c|}{\textbf{Token$_{\mu}$}} & \textbf{LR (\%)} \\
        & & \textbf{PFO} & \textbf{Unfil} & \\
        \midrule
        false & 594 & 589.77 & 788.05 & 25.16\% \\
        mostly-false & 600 & 808.06 & 1050.69 & 23.08\% \\
        half-true & 593 & 860.37 & 998.79 & 13.85\% \\
        mostly-true & 598 & 765.88 & 910.63 & 15.91\% \\
        true & 597 & 681.73 & 760.17 & 10.31\% \\
        \midrule
        \textbf{Total} & 2982 & 741.23 & 901.78 & 17.79\% \\
    \end{tabular}
    \caption{Summary statistics for each class label, including sample count (\textbf{Count}), average tokens per evidence in filtered (PFO) and unfiltered (Unfil) versions (\textbf{Token$_{\mu}$}), and percentage length reduction (\textbf{LR}) from Unfil to PFO. The final row shows totals and overall averages, with LR as the percent drop from overall unfiltered to filtered content.}
    \label{table:token_content_drop}
\end{table}
Our goal is to make the PFO dataset as representative as possible of \textit{real-world scenarios}, where no post-publication analysis (leakage) about the claim exists. To construct the PFO dataset, we began by scraping fact-checking articles from PolitiFact that correspond to claims in the PFO dataset. These articles typically include both factual information and annotator commentary that links various facts together. We manually removed any analysis that was introduced after the claim’s publication. Specifically, the annotator's interpretive or connective commentary. We retain only the facts that existed before the claim was published. Table~\ref{table:token_content_drop} provides a comparison of the \textit{Politi-Fact-Only} and its unfiltered version. We can see that on average \textbf{17.79\%} of the original dataset consisted of non-factual commentary.

% \begin{table*}[h]
% \centering
% \small
% \begin{tabular}{l|cc}
% \textbf{Dataset - Model} & \textbf{Respective Dataset} & \textbf{PolitiFact-Fact-Only} \\ 
% \midrule
% LIAR-PLUS - SVM \cite{alhindi-etal-2018-evidence}  & 0.25  & 0.27  \\
% LIAR - CNN \cite{wang-2017-liar}  & 0.27  & 0.28  \\
% LIAR-PLUS - LR \cite{alhindi-etal-2018-evidence}  & 0.37  & 0.27  \\
% % LIAR-RAW - CofCED \cite{yang2022cofced}  & 0.28  & —  \\
% % WATClaimCheck - RoBERTa-base \cite{khan-etal-2022-watclaimcheck}  & 0.58  & —  \\
% \text{{\large {AV}}}ERI\text{{\large {T}}}E\text{{\large {C}}} - BERT-large
%  \cite{schlichtkrull2023averitec}  & 0.49  & 0.29  \\
% \end{tabular}
% \caption{Comparison of \textit{Politi-Fact-Only} with other fact-checking datasets. All results are reported in Macro F1-score. The values for the respective datasets are sourced from the original authors' reported results.}
% \label{table:comparison_SOTA}
% \end{table*}

\subsection{Dataset filtration:}
\label{sec:annotation}
We removed sentences from the evidence if they were directly related to the verdict, such as statements like ``so the claim is incorrect''. Similarly, we eliminated annotator commentary, such as ``We found out that the claim is leaving out the partial information''. These logical or inference-based cues can lead to information leakage in an LM-based fact-checking system. However, in real-world scenarios, only factual information is available, making it harder to assess the claim’s veracity based solely on the provided evidence. Consequently, we removed 18 instances where, after the filtration process, the reduced context no longer provided enough information to determine the accuracy of the claim. We provide the information and statistics of source data, guidelines, and annotators information in Appendix ~\ref{sec:guidelines}.

We randomly sampled 200 instances from the PFO dataset to calculate the Inter Annotator Agreement of our three annotators and one annotator from PolitiFact who provided the gold label. We get a Fleiss' Kappa score of \textbf{0.7092}, indicating better agreement among the annotators.

\section{Recon-Answer-Verify (RAV)}
\label{sec:methodology}
We propose a multi-agent system, RAV, for claim verification. It decomposes a claim into sub-questions, then answers them using external context, and then predicts a final label. This is analogous to the real-world fact-checking process. The system comprises of three agents arranged in a pipeline: \textbf{QG\textsubscript{agent}} (Question Generator), \textbf{AG\textsubscript{agent}} (Answer Generator), and \textbf{LG\textsubscript{agent}} (Label Generator). These agents work collaboratively in an iterative process, where each cycle generates and answers a new question, ultimately leading to a final label for the claim (see Figure~\ref{fig:RAV}). The specific prompts used for each agent are provided in Appendix~\ref{methodology_prompt}. We name the pipeline RAV (Recon-Answer-Verify) because it conducts reconnaissance by iteratively generating questions and answers to investigate a given claim, followed by a final label prediction step. The specific roles of each agent involved are described in the following section. The complete RAV pipeline is outlined in Algorithm~\ref{alg:agentic_pipeline}.

\begin{figure*}
    \centering
    \includegraphics[width=0.9\linewidth]{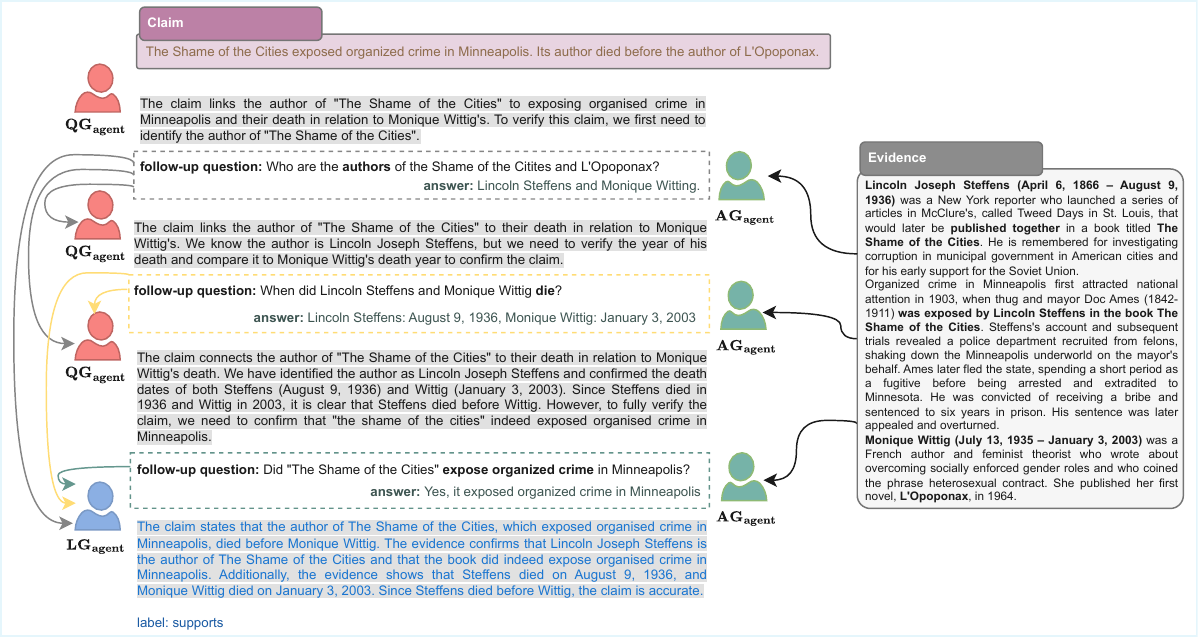}
    \caption{This diagram illustrates the workflow of the RAV pipeline for an instance (claim \& evidence pair), showcasing each stage from claim decomposition through question generation to reasoning, answer generation to final verdict generation. This instance required both True/False answerable questions (Q2 \& Q3) and inquiry-based questions (Q1), enabling iterative reasoning and complex claim verification. Text with a highlighted background in the workflow is the reasoning provided by the agent before generating the final output.}
    \label{fig:RAV}
\end{figure*}

\begin{algorithm}
\caption{RAV Pipeline for Claim Verification}
\label{alg:agentic_pipeline}
\begin{algorithmic}[1]
\REQUIRE Claim $C$, Evidence $E$
\STATE Initialize history $\mathcal{H}_0 \leftarrow \emptyset$, $t \leftarrow 1$
\WHILE{True}
    \STATE $(q_t, s_t, r_{QG}^*) \sim QG_{\text{agent}}(C, \mathcal{H}_{t-1})$
    \IF{$s_t = \texttt{true}$}
        \STATE \textbf{break}
    \ENDIF
    \STATE $a_t \sim AG_{\text{agent}}(q_t, E)$
    \STATE $\mathcal{H}_t \leftarrow \mathcal{H}_{t-1} \cup \{(q_t, a_t)\}$
    \STATE $t \leftarrow t + 1$
\ENDWHILE
\STATE $(y, r_{LG}^*) \sim LG_{\text{agent}}(C, \mathcal{H}_T)$
\RETURN Veracity label $y$ with reasoning $r_{LG}^*$
\end{algorithmic}
\end{algorithm}

\begin{itemize}[leftmargin=*]
    \item \textbf{QG\textsubscript{agent}:}  
    The Question Generation agent at timestep \(t\) produces a sub-question \( q_t \) based on the input claim \( C \), using the history of questions and answers up to step \( t-1 \), without access to external evidence. This enables domain-agnostic generalization. At each step, it generates a brief reasoning trace \( r_{QG}^* \) linking the claim with the history, which guides the question generation. The questions can be either verification (true/false answerable) or inquiry-based (requiring explanation). Verification question verifies a complete triple \(<entity_{1}, relationship, entity_{2}>\) and an inquiry question either enquires about the entity or the relationship. The agent also determines if the claim has been sufficiently explored by asking questions. It emits a stop signal \( s_t = \texttt{true} \) to end the process.

    \item \textbf{AG\textsubscript{agent}:}  
    The Answer Generation agent receives a generated question \( q_t \) and the external evidence \( E \) associated with the claim, and returns an answer \( a_t \). This step connects the verification question to the factual context needed to evaluate the claim.

    \item \textbf{LG\textsubscript{agent}:}  
    The Label Generation agent takes the original claim \( c \), along with the history of question-answer pairs \( \mathcal{H}_t = \{(q_1, a_1), \ldots, (q_t, a_t)\} \), and predicts the final veracity label \( y \) for the claim. It also produces a reasoning trace \( r_{LG}^* \) that explains how the label was derived based on the claim and the Q\&A history.
\end{itemize}

\section{Datasets}
\label{dataset_and_evaluation}
To evaluate and compare the proposed PFO dataset, we conduct zero-shot experiments using \texttt{Llama-3.1-8B}\footnote{\href{https://huggingface.co/meta-llama/Llama-3.1-8B}{huggingface.co/meta-llama/Llama-3.1-8B}}, \texttt{Mistral-7B-v0.3}\footnote{\href{https://huggingface.co/mistralai/Mistral-7B-v0.3}{huggingface.co/mistralai/Mistral-7B-v0.3}}, and \texttt{gemma-2-9b}\footnote{\href{https://huggingface.co/google/gemma-2-9b}{huggingface.co/google/gemma-2-9b}} on the PFO, and compare it with MultiFC, LIAR-PLUS, L++, RU22fact, and unfiltered versions of PFO. This setting evaluates the effect of evidence containing information that existed before the claim published versus evidence containing an annotator's commentary and post-publication information (see Table~\ref{table:model_comparison_macro}). Secondly, for experiments involving the RAV pipeline, we include 6 datasets with different domains and label granularities: RAWFC (3-class, multi-domain), LIAR-RAW, PFO and unfiltered (5-class, political), and HOVER and FEVEROUS (2-class, encyclopedic). Gold evidence is used for all datasets; in LIAR-RAW and RAWFC, we follow \citet{yang2022cofced} and treat author-written explanations as gold evidence, bypassing retrieval from source documents.

\section{Experimental Setup}
\label{experimental_setup}
We perform three types of experiments: (i) evaluation and comparison on the PFO dataset, (ii) evaluation and comparison of the RAV pipeline with various baselines, and (iii) ablation experiments for the RAV pipeline. 

\begin{figure}
    \centering
    \includegraphics[width=0.9\linewidth]{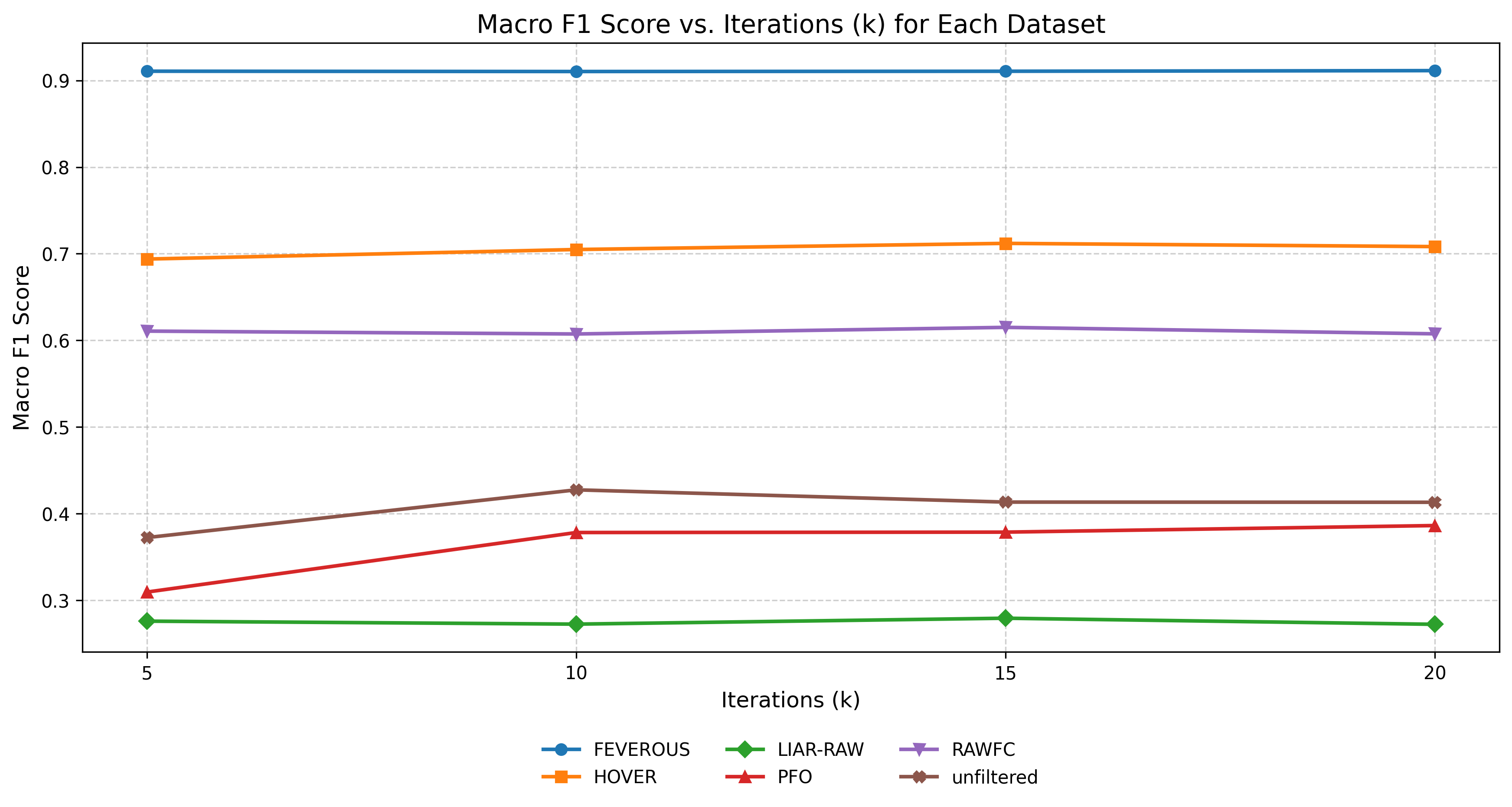}
    \caption{Macro F1 score of the RAV pipeline across six datasets as a function of the maximum allowed number of QA iterations (k). Each iteration represents an additional step in the agentic reasoning process, with the option for early stopping when all aspects of the claim are covered by the questions. We observe that performance tends to plateau by iteration 10 across most datasets, indicating that further iterations yield diminishing returns. Based on this saturation trend and to balance performance with computational efficiency, we select 10 iterations as the default setting for our final experiments.}
    \label{fig:itr_vs_macrof1}
\end{figure}

% \subsection{Implementation Details}
% \label{sec:implementation_details}
To select the best prompt for the zero-shot setting, we evaluated seven prompt variants across three language models: Mistral-7B-v0.3, Llama-3-8B, and Gemma-2-9B. As shown in Table~\ref{table:promptselectionresults} in Appendix~\ref{sec:promptselection}, even slight changes in phrasing caused notable shifts in F1 scores—up to 10 points in some cases. Details of the prompt variants are provided in Appendix~\ref{sec:zeroshot}. \emph{P6} consistently achieved the highest performance across models and was used in all subsequent zero-shot experiments. We conduct experiments varying the number of reasoning iterations (k) across values {5, 10, 15, 20}, and observe that performance plateaus or degrades beyond k = 10 (see Figure~\ref{fig:itr_vs_macrof1}). Therefore, we set a maximum limit of 10 verification questions per instance. Models used include \texttt{Mistral-7B-Instruct-v0.3}\footnote{\href{https://huggingface.co/mistralai/Mistral-7B-Instruct-v0.3}{huggingface.co/mistralai/Mistral-7B-Instruct-v0.3}}, \texttt{Llama-3.1-8B-Instruct}\footnote{\href{https://huggingface.co/meta-llama/Llama-3.1-8B-Instruct}{huggingface.co/meta-llama/Llama-3.1-8B-Instruct}}, \texttt{gemma-2-9b-it}\footnote{\href{https://huggingface.co/google/gemma-2-9b-it}{huggingface.co/google/gemma-2-9b-it}}, \texttt{phi-4}\footnote{\href{https://huggingface.co/microsoft/phi-4}{huggingface.co/microsoft/phi-4}}, and \texttt{Llama-3.1-70B-Instruct}\footnote{\href{https://huggingface.co/meta-llama/Llama-3.1-70B-Instruct}{huggingface.co/meta-llama/Llama-3.1-70B-Instruct}} (see Table~\ref{tab:main_comparison_variant4}). For ablation, we test four RAV variants that differ in question generation strategy (all-at-once vs. iterative) and question type (True/False vs. mixed inquiry), using LIAR-RAW, RAWFC, and FEVEROUS with \texttt{phi-4} and \texttt{Llama-3.1-70B-Instruct}. The best-performing setup, RAV(\(P_2, T_{1\&2}\)), highlights the benefit of iterative and diverse questioning (see Table~\ref{tab:ablation_models_first}). Further methodological details are in Section~\ref{sec:methodology}.

\subsection{Baselines}
\label{sec:baselines}
We consider three existing fact-checking approaches: ProgramFC \cite{pan2023factchecking}, CofCED \cite{yang2022cofced}, and HiSS \cite{zhang2023llmbased}, as baselines. For baselines, HiSS and CofCed, we directly report results reported by authors on standard benchmark datasets. CofCED is a supervised technique for the fact-checking. However, ProgramFC and Hiss use a 175B (\texttt{text-davinci-003}) as the LLM's backbone.

\begin{table*}[t]
\centering
\small
\begin{tabular}{l|ccccccc}
\textbf{Models} & \textbf{MultiFC} & \textbf{LIAR-PLUS} & \textbf{RU22fact} & \textbf{L++} & \textbf{PFO} & \textbf{Unfiltered} \\ 
\midrule
Mistral-7b-v0.3     & 0.14/\textbf{0.29} & 0.14/\textbf{0.29}  & 0.32/\textbf{0.65}  & 0.28/\textbf{0.36}  & 0.26/\textbf{0.34}  & 0.37/\textbf{0.45}  \\
LLaMA-3.1-8b     & 0.15/\textbf{0.25} & 0.15/\textbf{0.25}  & 0.29/\textbf{0.64}  & 0.28/\textbf{0.35}  & 0.21/\textbf{0.28}  & 0.48/\textbf{0.51}  \\
Gemma-2-9b    & 0.15/\textbf{0.25} & 0.15/\textbf{0.25}  & 0.27/\textbf{0.64}  & 0.18/\textbf{0.31}  & 0.20/\textbf{0.28}  & 0.59/\textbf{0.60}  \\
\end{tabular}
\caption{Performance comparison of models ranging from 7B to 9B parameters using Zero-Shot prompting \cite{10.5555/3600270.3601883} across various fact-checking datasets. The results are reported in macro-F1/micro-F1 score. The "Unfiltered" dataset represents the unfiltered version of \textit{Politi-Fact-Only}. We use models like \texttt{meta-llama/Meta-Llama-3.1-8B}, \texttt{mistralai}
\texttt{/Mistral-7B-v0.3},  a model from Google from GEMMA series \texttt{google/gemma-2-9b} from Huggingface.}
\label{table:model_comparison_macro}
\end{table*}

\subsection{Ablation}
\label{subsec:varients}
We experiment with combinations of two types of question generation strategies and two questioning types for the Question Generator. These strategies affect how questions are generated, and indirectly, how the final label is decided. 

We consider two different strategies for generating the questions. \textbf{All questions at once} (\(\mathbf{P_{1}}\)): The \(QG_{agent}\) starts with a reasoning to explore different sub-parts of the claim, then all questions needed to verify the claim are generated at once (\(S_{1}\)). The \(AG_{agent}\) generates answers for all questions at once (\(S_{2}\)). The \(LG_{agent}\) then generates a reasoning that connects the claim, the questions and the answers, after which it predicts the label (\(S_{3}\)). \textbf{Iterative question generation} (\(\mathbf{P_{2}}\)): It is an iterative process with \(S_{1}\) and \(S_{2}\) of \(\mathbf{P_{1}}\) in a loop. The \(QG_{agent}\) generates questions one at a time in an iterative loop with \(AG_{agent}\). After each iteration, the \(QG_{agent}\) decides whether to generate another question or stop. The iterative process stops when the \(QG_{agent}\) reasons that no further questions are needed to verify the claim. \textit{Step 3} is similar to \(\mathbf{P_{1}}\).

To simplify the process of claim decomposition, we generate either a \emph{Verification} question (true/false answerable), which confirms a complete triple  \(<\mathrm{entity}_{1}, \mathrm{relationship}, \mathrm{entity}_{2}>\) or \emph{Inquiry questions}, requiring an entity or a relationship in the response. We consider two types of questioning \(T_{1}\) and \(T_{2}\). \(\mathbf{T_{1}}\): A True/False answerable question that verifies a complete triple. For example, \textit{``Did The Sham of the Cities expose organized crime in Minneapolis?''}, here it verifies a triple \texttt{<The Sham of the Cities, expose organized crime, Minneapolis>}. \(\mathbf{T_{2}}\): An Inquiry Question that either requires entities related to the other entity, or the relationship as a response. For example, \textit{``Who are the authors of the Shame of the Cities and L'Opoponax?''} and \textit{``When did Lincoln Steffens and Monique Witting die?''}. Both questions are asking for the related entities in the response.

To denote all the variants cleraly, we assume a function RAV(\(P_{t}\), \(Q_{t}\)), which takes a question generation strategy \(P_{t}\) and a questioning type \(Q_{t}\) as input and follows the process accordingly. Our 4 variants are: \textbf{(1)} RAV(\(P_{1}\), \(T_{1}\)), \textbf{(2)} RAV(\(P_{2}\), \(T_{1}\)), \textbf{(3)} RAV(\(P_{1}\), \(T_{1\&2}\)), \textbf{(4)} RAV(\(P_{2}\), \(T_{1\&2}\)), where fourth variants is proposed RAV pipeline. For variants (1) and (3), we generate 3 question trajectories for an instance, and after the label prediction, we do the majority voting and select the label.

\begin{table*}[t]
\centering
\small
\begin{tabular}{c|cccccccc}
\toprule
\textbf{Model / Macro-F1} & \textbf{PFO} & \textbf{Unfiltered} & \textbf{LIAR-RAW} & \textbf{RAWFC} & \multicolumn{3}{c}{\textbf{Hover}} & \textbf{Fever} \\
 & & & & & \textbf{2-hop} & \textbf{3-hop} & \textbf{4-hop} & \\
\midrule
ProgramFC         & -- & -- & -- & -- & 0.7565\textsuperscript{*} & 0.6848\textsuperscript{*} & 0.6675\textsuperscript{*} & \textbf{0.9269}\textsuperscript{*} \\
CofCED            & -- & -- & 0.2893 & 0.5107 & -- & -- & -- & -- \\
HiSS            & -- & -- & \textbf{0.3750}\textsuperscript{*} & 0.5390\textsuperscript{*} & -- & -- & -- & -- \\
\midrule
RAV (Mistral-7b-v0.3)  & 0.2724 & 0.3660 & 0.2169 & 0.4916 & 0.6576 & 0.5120 & 0.4609 & 0.8098 \\
RAV (LLaMA-3.1-8b) & 0.2678 & 0.3282 & 0.2123 & 0.4893 & 0.6680 & 0.5868 & 0.5827 & 0.7720 \\
RAV (Gemma-2-9b)  & 0.1642 & 0.2391 & 0.1265 & 0.3512 & 0.7067 & 0.6240 & 0.5969 & 0.8582 \\
RAV (phi-4) & 0.3701 & 0.4235 & 0.2933 & \textbf{0.6753} & 0.7558 & 0.6753 & 0.6486 & 0.9121 \\
RAV (LLaMA-3.1-70b) & \textbf{0.3768} & \textbf{0.4504} & 0.2540 & 0.5919 & \textbf{0.7682} & \textbf{0.7188} & \textbf{0.6794} & 0.9063 \\
\midrule
$\text{RAV}^{p}$ (\text{LLaMA-3.1-70b}) & 0.2649 & 0.2243 & 0.2373 & 0.4953 & 0.6465 & 0.6109 & 0.5897 & 0.6293 \\
\bottomrule
\end{tabular}
\caption{Macro-F1 scores of three baseline models, ProgramFC \cite{pan2023factchecking}, HiSS \cite{zhang2023llmbased}, and CofCED \cite{yang2022cofced}, as well as our proposed agentic RAV (Recon-Answer-Verify) pipeline, evaluated using multiple instruct LLMs (\texttt{Mistral-7B-Instruct-v0.3}, \texttt{Llama-3.1-8B/70B-Instruct}, \texttt{phi-4} and \texttt{Gemma-2-9B-it}). Evaluations are conducted across multiple fact verification datasets. Here, "Unfiltered" dataset represents the unfiltered version of \textit{Politi-Fact-Only} (PFO). * mark on the number shows the results are reported using the GPT-3.5 series 175B parameters model \texttt{text-davinci-003}\footnote{\url{https://platform.openai.com/docs/models}}. $\text{RAV}^{p}$ (\text{LLaMA-3.1-70b}) denotes that for the \(AG_{agent}\), the RAV pipeline leverages only the pretrained knowledge of its backbone LLM (LLaMA-3.1-70b-Instruct) to answer the question.
}
\label{tab:main_comparison_variant4}
\end{table*}

\begin{table*}[t]
\centering
\small
\resizebox{\textwidth}{!}{%
\begin{tabular}{c|cccc|cccc}
\toprule
\textbf{Variant} 
& \multicolumn{4}{c|}{\textbf{phi-4}} 
& \multicolumn{4}{c}{\textbf{llama-3.1-70B-Instruct}} \\
& \textbf{LIAR-RAW} & \textbf{RAWFC} & \textbf{Fever} & \textbf{Hover}
& \textbf{LIAR-RAW} & \textbf{RAWFC} & \textbf{Fever} & \textbf{Hover} \\
\midrule
\(RAV_{E}\)(\(P_{2}, T_{1\&2}\)) & 0.3215 & 0.5341 & 0.8878 & 0.6669 & 0.3469 & 0.6184 & 0.8860 & 0.6881 \\
\midrule
RAV(\(P_{1}, T_{1}\))           & 0.3002 & 0.5870 & 0.8811 & 0.6671   & 0.3210 & 0.6814 & 0.8608 & 0.6975      \\
RAV(\(P_{2}, T_{1}\))           & 0.2893 & 0.6112 & 0.8960 & 0.7156 & 0.2491 & 0.5953 & 0.8842 & 0.7043 \\
RAV(\(P_{1}, T_{1\&2}\))        & \textbf{0.3279} & 0.6390 & 0.8845 & 0.6825   & \textbf{0.3629} & \textbf{0.7169} & 0.8772 & 0.6840      \\
RAV(\(P_{2}, T_{1\&2}\)) (ours) & 0.2933 & \textbf{0.6753} & \textbf{0.9121} & \textbf{0.7228} & 0.2540 & 0.5919 & \textbf{0.9063} & \textbf{0.7227} \\
\bottomrule
\end{tabular}%
}
\caption{Macro-F1 scores of four RAV pipeline variants evaluated on LIAR-RAW, RAWFC, FEVEROUS, and HOVER datasets using two backbone models: \texttt{microsoft/phi-4} (14B parameters) and \texttt{meta-llama/Llama-3.1-70B-Instruct} (70B parameters). The variants differ in Question Generator design and question types used (see Section~\ref{subsec:varients}). Variant 4, which uses the full agentic pipeline described in Section~\ref{sec:methodology}, achieves the highest scores on most datasets, demonstrating the benefit of iterative and diverse question generation in claim verification. Missing values indicate the variant was not evaluated on that dataset.}
\label{tab:ablation_models_first}
\end{table*}

\begin{table}[h!]
\small
\begin{tabular}{@{}lcccc@{}}
\toprule
\textbf{Dataset} & \multicolumn{2}{c}{\textbf{phi-4}} & \multicolumn{2}{c}{\textbf{llama-3.1-70b-instruct}} \\
\cmidrule(lr){2-3} \cmidrule(lr){4-5}
& \textit{w/o} res & \textit{w/} res & \textit{w/o} res & \textit{w/} res \\
\midrule
FEVEROUS & 0.8914 & \textbf{0.9121} & 0.8744 & \textbf{0.9063} \\
RAWFC    & 0.5980 & \textbf{0.6753} & 0.5545 & \textbf{0.5919} \\
PFO      & 0.3816 & 0.3701 & 0.3457 & \textbf{0.3768} \\
\bottomrule
\end{tabular}
\caption{Macro F1 scores of two language models, \texttt{phi-4} and \texttt{llama-3.1-70b-instruct}, evaluated on the RAV pipeline across three fact-checking datasets: FEVEROUS, RAWFC, and PFO. Scores are reported under two prompting conditions: without reasoning (w/o res) and with reasoning (w/ res). The results demonstrate that incorporating explicit reasoning prompts leads to performance improvements on FEVEROUS and RAWFC, while performance on PFO remains relatively stable.}
\label{tab:woreasoning}
\end{table}

\section{Results and Error Analysis}
\label{sec:experiments_and_result}
We first evaluated 3 LLMs \texttt{Meta-LLaMA-3.1-8B}, \texttt{Mistral-7B-v0.3}, and \texttt{Gemma-2-9b} on multiple fact-checking datasets, including \textit{LIAR-PLUS}, \textit{RU22fact}, \textit{L++}, and \textit{PFO} and unfiltered PFO. Since LLM performance is highly sensitive to prompt design, we conducted a prompt optimization experiment to identify the most effective prompt for final evaluations. Appendix Table \ref{table:promptselectionresults} illustrates that even a single keyword change in prompt can impact model outputs. Our findings emphasize the need for prompt engineering when leveraging LLMs. As shown in Table \ref{table:model_comparison_macro}, we report macro and micro F1-scores across different datasets. Among the evaluated models, \textit{Mistral-7B-v0.3} consistently achieved the highest overall F1-score, highlighting its effectiveness in zero-shot fact verification. Notably, LLMs, when evaluated on the unfiltered version of \textit{PFO} (referred to as Unfiltered), show increased and false performance, suggesting that implicit cues or post-analysis commentary contribute to inflated model performance. LLMs, when evaluated on \textit{LIAR-PLUS}, scored lower than \textit{PFO}, likely because approximately 50\% of their dataset includes evidence derived from only the last five lines of the article \cite{10.1162/tacl_a_00601}, leading to weaker performance due to insufficient evidence. For multifc, the claim is submitted verbatim as a query to the Google Search API (without quotes). The 10 most highly ranked search results are retrieved, but the top 10 do not ensure the complete information needed to detect the veracity of the claim.

\paragraph{Comparing Fact-checking systems.}We compare various state-of-the-art fact-checking systems (Section~\ref{sec:baselines}) with our RAV pipeline. We run the RAV pipeline with various LLM backbones. On all the benchmarks except Feverous, we see that RAV outperfoms the baseline approaches. Compared to ProgramFC we observe an average increase in macro-F1 score by $\mathbf{2.1\%}$. Similarly, for the CofCED and HiSS baselines, we see an increase of macro-F1 score by $\mathbf{32.23\%}$ and $\mathbf{25.28\%}$ respectively. We also observe that \texttt{phi-4} and \texttt{LlaMA-3.1-70b} backbones show the best performance compared to other LLM backbones in RAV.

\paragraph{Robustness of RAV.}RAV demonstrates robustness when evaluated on both the PFO dataset and its unfiltered variant. On average, RAV experiences a 16.3\% drop in macro-F1 score. However, our best-performing RAV model, built on the LLaMA-3.1-70B backbone, shows a significantly smaller performance drop of just 7.36\%, compared to a much larger 22\% average drop observed in zero-shot settings.

\paragraph{Ablation Study.}RAV pipeline consistently outperforms the other variations. Compared to RAV(\(P_{1}\), \(T_{1}\)) we see an average performance difference of $\mathbf{5.43\%}$. This highlights the importance of the iterative process of question-answering employed by RAV pipelines. Compared to RAV(\(P_{2}\), \(T_{1}\)) we see an average performance difference of $\mathbf{11.82\%}$. As RAV asks both verification and inquiry questions, it is able to decompose the claim into sub-questions significantly better. Compared to RAV(\(P_{1}\), \(T_{1\&2}\)) we see an average performance difference of $\mathbf{1.69\%}$. As the iterative pipeline allows generation of subquestions conditioned on the previous answers (along with the previous questions and claim), the subquestions are able to capture all aspects of the claim. This signifies that the overall design of the RAV pipeline effectively addresses the challenges of veracity detection.

\paragraph{RAV with and without reasoning.}To analyze the effect of reasoning steps in the RAV pipeline (during question generation and label prediction), we compare RAV with and without the use of reasoning steps. Table~\ref{tab:woreasoning} presents the results of comparing the two. Over the datasets of 2-class, 3-class and 5-class labels, without the reasoning steps, RAV shows an average performance degradation of $\mathbf{3.11\%}$. This highlights the significance of the reasoning steps during the fact-checking process.

\paragraph{Analyzing reasoning complexity.}Benchmark datasets Hover and Feverous categorize the claims based on their complexity. Hover consists of claims belonging to 2-hop, 3-hop and 4-hop categories. Feverous categorizes claims based on the specific challenge as: Search terms not in claim, Entity Disambiguation, Multi-hop Reasoning, Numerical Reasoning, Combining Tables and Text, and Other. We quantify the reasoning complexity of a category as the number of subquestions a claim belonging to each category is broken into. For the Hover dataset, we observe that claims belonging to the 2-hop class require 4 sub-questions on average, the 3-hop class is broken into 5, and the 4-hop class is broken into 6 sub-questions on average. These observations underline the increasing requirement for reasoning capability of fact-checking systems as the claim complexity increases. Similarly, for the Feverous dataset, claims belonging to the Combining Tables and Text category are broken into 8 questions on average, and claims belonging to Numerical Reasoning are broken into 7 questions on average. Other categories are broken down into 5-6 sub-questions. This indicates that the former two categories are more complex than the latter. We show the trend of the reasoning complexity in Appendix~\ref{app:rav_pipeline_analysis}.

\noindent We show the human analysis of output from the RAV pipeline in Appendix~\ref{app:human_evaluation_of_rav}.

\section{Conclusion and Future Work}
In this work we contribute, \textit{PFO}, a benchmark dataset for evaluating fact-checking models using only factual evidence, without post-claim analysis and \textit{RAV} pipeline, a fact-checking system which can handle claims of multiple domains and label granularities. Our experiments show that models struggle when deprived of annotator cues, resulting in a performance drop compared to unfiltered datasets. This highlights LLM's reliance on post-claim analysis provided by fact-checkers. We experiment with different variants of the question generation strategies and find out that our final pipeline performs better than the other variants. These contributions towards automating the fact-checking process of real-world claims. Through qualitative analysis, we observe that the answer generator suffers due to insufficient evidence. As we conduct our experiments with gold evidence provided in the evaluation datasets, for future work, we aim to utilise the external context as evidence, and if we imply external context as evidence, then we need to use the retriever. Further future work can include adapting retrievers for the question answering process in our pipeline. 

\section*{Limitation}
This dataset is collected from a fact-checking website. While we have attempted to remove most annotator cues, some sentences could not be eliminated without compromising the context necessary to support or refute the claim.

% \section*{Acknowledgements}

\label{sec:bibtex}
\bibliography{custom}

\begin{thebibliography}{19}
\expandafter\ifx\csname natexlab\endcsname\relax\def\natexlab#1{#1}\fi

\bibitem[{Alhindi et~al.(2018)Alhindi, Petridis, and Muresan}]{alhindi-etal-2018-evidence}
Tariq Alhindi, Savvas Petridis, and Smaranda Muresan. 2018.
\newblock \href {https://doi.org/10.18653/v1/W18-5513} {Where is your evidence: Improving fact-checking by justification modeling}.
\newblock In \emph{Proceedings of the First Workshop on Fact Extraction and {VER}ification ({FEVER})}, pages 85--90, Brussels, Belgium. Association for Computational Linguistics.

\bibitem[{Augenstein et~al.(2019)Augenstein, Lioma, Wang, Chaves~Lima, Hansen, Hansen, and Simonsen}]{augenstein-etal-2019-multifc}
Isabelle Augenstein, Christina Lioma, Dongsheng Wang, Lucas Chaves~Lima, Casper Hansen, Christian Hansen, and Jakob~Grue Simonsen. 2019.
\newblock \href {https://doi.org/10.18653/v1/D19-1475} {{M}ulti{FC}: A real-world multi-domain dataset for evidence-based fact checking of claims}.
\newblock In \emph{Proceedings of the 2019 Conference on Empirical Methods in Natural Language Processing and the 9th International Joint Conference on Natural Language Processing (EMNLP-IJCNLP)}, pages 4685--4697, Hong Kong, China. Association for Computational Linguistics.

\bibitem[{Cohen et~al.(2023)Cohen, Hamri, Geva, and Globerson}]{cohen2023lm}
Roi Cohen, May Hamri, Mor Geva, and Amir Globerson. 2023.
\newblock Lm vs lm: Detecting factual errors via cross examination.
\newblock \emph{arXiv preprint arXiv:2305.13281}.

\bibitem[{Graves(2018)}]{graves2018understanding}
Lucas Graves. 2018.
\newblock Understanding the promise and limits of automated fact-checking.
\newblock \emph{Reuters Institute for the Study of Journalism}.

\bibitem[{Gupta and Srikumar(2021)}]{gupta-srikumar-2021-x}
Ashim Gupta and Vivek Srikumar. 2021.
\newblock \href {https://doi.org/10.18653/v1/2021.acl-short.86} {{X}-fact: A new benchmark dataset for multilingual fact checking}.
\newblock In \emph{Proceedings of the 59th Annual Meeting of the Association for Computational Linguistics and the 11th International Joint Conference on Natural Language Processing (Volume 2: Short Papers)}, pages 675--682, Online. Association for Computational Linguistics.

\bibitem[{Jiang et~al.(2020)Jiang, Bordia, Zhong, Dognin, Singh, and Bansal}]{jiang-etal-2020-hover}
Yichen Jiang, Shikha Bordia, Zheng Zhong, Charles Dognin, Maneesh Singh, and Mohit Bansal. 2020.
\newblock \href {https://doi.org/10.18653/v1/2020.findings-emnlp.309} {{H}o{V}er: A dataset for many-hop fact extraction and claim verification}.
\newblock In \emph{Findings of the Association for Computational Linguistics: EMNLP 2020}, pages 3441--3460, Online. Association for Computational Linguistics.

\bibitem[{Khan et~al.(2022)Khan, Wang, and Poupart}]{khan-etal-2022-watclaimcheck}
Kashif Khan, Ruizhe Wang, and Pascal Poupart. 2022.
\newblock \href {https://doi.org/10.18653/v1/2022.acl-long.92} {{W}at{C}laim{C}heck: A new dataset for claim entailment and inference}.
\newblock In \emph{Proceedings of the 60th Annual Meeting of the Association for Computational Linguistics (Volume 1: Long Papers)}, pages 1293--1304, Dublin, Ireland. Association for Computational Linguistics.

\bibitem[{Kojima et~al.(2024)Kojima, Gu, Reid, Matsuo, and Iwasawa}]{10.5555/3600270.3601883}
Takeshi Kojima, Shixiang~Shane Gu, Machel Reid, Yutaka Matsuo, and Yusuke Iwasawa. 2024.
\newblock Large language models are zero-shot reasoners.
\newblock In \emph{Proceedings of the 36th International Conference on Neural Information Processing Systems}, NIPS '22, Red Hook, NY, USA. Curran Associates Inc.

\bibitem[{Lewis et~al.(2008)Lewis, Williams, Franklin, Thomas, and Mosdell}]{lewis2008quality}
Justin Matthew~Wren Lewis, Andy Williams, Robert~Arthur Franklin, James Thomas, and Nicholas~Alexander Mosdell. 2008.
\newblock The quality and independence of british journalism.

\bibitem[{Misra(2022)}]{misra2022politifact}
Rishabh Misra. 2022.
\newblock \href {https://doi.org/10.13140/RG.2.2.29923.22566} {Politifact fact check dataset}.

\bibitem[{Pan et~al.(2023)Pan, Wu, Lu, Luu, Wang, Kan, and Nakov}]{pan2023factchecking}
Liangming Pan, Xiaobao Wu, Xinyuan Lu, Anh~Tuan Luu, William~Yang Wang, Min-Yen Kan, and Preslav Nakov. 2023.
\newblock \href {https://arxiv.org/abs/2305.12744} {Fact-checking complex claims with program-guided reasoning}.

\bibitem[{Rashkin et~al.(2017)Rashkin, Choi, Jang, Volkova, and Choi}]{rashkin-etal-2017-truth}
Hannah Rashkin, Eunsol Choi, Jin~Yea Jang, Svitlana Volkova, and Yejin Choi. 2017.
\newblock \href {https://doi.org/10.18653/v1/D17-1317} {Truth of varying shades: Analyzing language in fake news and political fact-checking}.
\newblock In \emph{Proceedings of the 2017 Conference on Empirical Methods in Natural Language Processing}, pages 2931--2937, Copenhagen, Denmark. Association for Computational Linguistics.

\bibitem[{Russo et~al.(2023)Russo, Tekiroğlu, and Guerini}]{10.1162/tacl_a_00601}
Daniel Russo, Serra~Sinem Tekiroğlu, and Marco Guerini. 2023.
\newblock \href {https://doi.org/10.1162/tacl_a_00601} {{Benchmarking the Generation of Fact Checking Explanations}}.
\newblock \emph{Transactions of the Association for Computational Linguistics}, 11:1250--1264.

\bibitem[{Thorne et~al.(2018)Thorne, Vlachos, Christodoulopoulos, and Mittal}]{thorne-etal-2018-fever}
James Thorne, Andreas Vlachos, Christos Christodoulopoulos, and Arpit Mittal. 2018.
\newblock \href {https://doi.org/10.18653/v1/N18-1074} {{FEVER}: a large-scale dataset for fact extraction and {VER}ification}.
\newblock In \emph{Proceedings of the 2018 Conference of the North {A}merican Chapter of the Association for Computational Linguistics: Human Language Technologies, Volume 1 (Long Papers)}, pages 809--819, New Orleans, Louisiana. Association for Computational Linguistics.

\bibitem[{Vosoughi et~al.(2018)Vosoughi, Roy, and Aral}]{vosoughi2018spread}
Soroush Vosoughi, Deb Roy, and Sinan Aral. 2018.
\newblock The spread of true and false news online.
\newblock \emph{Science}, 359(6380):1146--1151.

\bibitem[{Wang(2017)}]{wang-2017-liar}
William~Yang Wang. 2017.
\newblock \href {https://doi.org/10.18653/v1/P17-2067} {{``}liar, liar pants on fire{''}: A new benchmark dataset for fake news detection}.
\newblock In \emph{Proceedings of the 55th Annual Meeting of the Association for Computational Linguistics (Volume 2: Short Papers)}, pages 422--426, Vancouver, Canada. Association for Computational Linguistics.

\bibitem[{Yang et~al.(2022)Yang, Ma, Chen, Lin, Luo, and Yi}]{yang2022cofced}
Zhiwei Yang, Jing Ma, Hechang Chen, Hongzhan Lin, Ziyang Luo, and Chang Yi. 2022.
\newblock \href {https://aclanthology.org/2022.coling-1.230} {A coarse-to-fine cascaded evidence-distillation neural network for explainable fake news detection}.
\newblock In \emph{Proceedings of the 29th International Conference on Computational Linguistics (COLING)}, pages 2608--2621.

\bibitem[{Zeng et~al.(2024)Zeng, Ding, Zhao, Li, Zhang, Yao, Liu, and Qin}]{zeng-etal-2024-ru22fact}
Yirong Zeng, Xiao Ding, Yi~Zhao, Xiangyu Li, Jie Zhang, Chao Yao, Ting Liu, and Bing Qin. 2024.
\newblock \href {https://aclanthology.org/2024.lrec-main.1239/} {{RU}22{F}act: Optimizing evidence for multilingual explainable fact-checking on {R}ussia-{U}kraine conflict}.
\newblock In \emph{Proceedings of the 2024 Joint International Conference on Computational Linguistics, Language Resources and Evaluation (LREC-COLING 2024)}, pages 14215--14226, Torino, Italia. ELRA and ICCL.

\bibitem[{Zhang and Gao(2023)}]{zhang2023llmbased}
Xuan Zhang and Wei Gao. 2023.
\newblock \href {http://arxiv.org/abs/2310.00305} {Towards llm-based fact verification on news claims with a hierarchical step-by-step prompting method}.

\end{thebibliography}
\bibliographystyle{acl_natbib}

% \clearpage
\appendix
\section{Prompt Selection}
\label{sec:promptselection}
As discussed in Section~\ref{experimental_setup}, in this section, we present the various prompts explored to identify the most effective one for the 5-class fact-checking task. We also report micro F1 scores in Table \ref{table:promptselectionresults} for each prompt evaluated on the validation set, providing insight into the performance differences across the prompt variations.
\subsection{Zero-Shot Base Model Prompts}
\label{sec:zeroshot}
In this section, we provide the seven prompts used for the base model in the zero-shot setting for the 5-class fact-checking task.
\begin{enumerate}[label=P\arabic*]
    \item \texttt{Given claim and evidence, predict if the claim is true, mostly-true, half-true, mostly-false, or false.}\\
    \texttt{claim: \{\{claim\}\}}\\
    \texttt{evidence: \{\{evidence\}\}}\\
    \texttt{label:}
    
    \item \texttt{Given the evidence, decide if the given claim is true, mostly-true, half-true, mostly-false, or false.}\\
    \texttt{claim: \{\{claim\}\}}\\
    \texttt{evidence: \{\{evidence\}\}}\\
    \texttt{label:}
    
    \item \texttt{Given claim and evidence, find if the claim is true, mostly-true, half-true, mostly-false, or false.}\\
    \texttt{claim: \{\{claim\}\}\ }\\
    \texttt{evidence: \{\{evidence\}\}}\\
    \texttt{label:}

    \item \texttt{Identify if the claim is true, mostly-true, half-true, mostly-false, or false based on the evidence.}\\
    \texttt{claim: \{\{claim\}\}}\\
    \texttt{evidence: \{\{evidence\}\}}\\
    \texttt{label:}
    
    \item \texttt{Given claim and evidence, classify if the claim is true, mostly-true, half-true, mostly-false, or false.}\\
    \texttt{claim: \{\{claim\}\}}\\
    \texttt{evidence: \{\{evidence\}\}}\\
    \texttt{label:}

    \item \texttt{You need to determine the accuracy of a claim based on the evidence. Use one of following 5 labels for the claim: true, mostly-true, half-true, mostly-false, or false. Examine the evidence and choose the most likely label based on the claim's accuracy without explaining your reasoning.}\\
    \texttt{claim: \{\{claim\}\}}\\
    \texttt{evidence: \{\{evidence\}\}}\\
    \texttt{label:}

    \item \texttt{Given claim and evidence, you are tasked with evaluating the truthfulness of claims based on the provided evidence. Each claim can be categorized into one of 5 labels: true, mostly-true, half-true, mostly-false, false. Assess the claim given the evidence and classify it appropriately without providing an explanation.}\\
    \texttt{claim: \{\{claim\}\}}\\
    \texttt{evidence: \{\{evidence\}\}}\\
    \texttt{label:}
\end{enumerate}

\begin{table*}[!]
\centering
\small
\resizebox{\textwidth}{!}{%
\begin{tabular}{llllllll}
\toprule
\multicolumn{8}{c}{\textbf{Zero Shot}}  \\ \midrule
\textbf{}                                              & \textbf{P1}     & \textbf{P2} & \textbf{P3} & \textbf{P4}     & \textbf{P5}     & \textbf{P6}     & \textbf{P7}     \\ \midrule                                             
\multicolumn{1}{l|}{\textbf{Mistral-7B-v0.3}}          & 0.3213          & 0.3213      & 0.3199      & 0.3396          & 0.3415          & \textbf{0.4253} & 0.4147          \\
\multicolumn{1}{l|}{\textbf{Llama-3-8B}}               & 0.2900            & 0.4607      & 0.4891      & 0.4678          & 0.4468          & \textbf{0.5202} & 0.4781          \\
\multicolumn{1}{l|}{\textbf{Gemma-2-9b}}               & 0.2979          & 0.3180      & 0.3264      & 0.3494          & 0.3094          & 0.3473          & \textbf{0.3769} \\ \bottomrule
\end{tabular}%
}
\caption{F1 Scores using the unfiltered version of Politi-Fact-Only dataset across various models using different prompt configurations with the Zero-Shot technique on the validation set. The results demonstrate how performance varies with different prompt selections, helping to identify the most effective prompt for the task.}
\label{table:promptselectionresults}
\end{table*}

\section{Annotators Information and Guidelines}
\label{sec:guidelines}
During the annotation time, we employed three annotators who were proficient in English and were compensated by us. They were paid 20,000 INR for this task each. On average, they took 20-25 minutes to clean an instance of the dataset, so it took around 3 months to clean the dataset. As we discussed the filtration process of the PFO dataset in Section~\ref{sec:annotation}. The following descriptions were given to the annotators related to the dataset.
\begin{table*}[t!]
\centering
\small
\resizebox{\textwidth}{!}{%
\begin{tabular}{lccc|c}
\toprule
\multicolumn{1}{c|}{\textbf{Label}} & \multicolumn{1}{c}{\textbf{Train Instances}} & \multicolumn{1}{c}{\textbf{Test Instances}} & \multicolumn{1}{c|}{\textbf{Validation Instances}} & \multicolumn{1}{c}{\textbf{Total Instances}} \\ \midrule
\multicolumn{1}{l|}{\textbf{True}}                      & 1,717                                        & 612                                         & 125                                                & 2,454                                        \\
\multicolumn{1}{l|}{\textbf{Mostly True}}               & 2,332                                        & 824                                         & 169                                                & 3,325                                        \\
\multicolumn{1}{l|}{\textbf{Half True}}                 & 2,502                                        & 910                                         & 179                                                & 3,591                                        \\
\multicolumn{1}{l|}{\textbf{Mostly False}}              & 2,409                                        & 829                                         & 184                                                & 3,422                                        \\
\multicolumn{1}{l|}{\textbf{False}}                     & 5,811                                        & 2,101                                       & 398                                                & 8,310                                        \\ \midrule
\multicolumn{1}{l|}{\textbf{Total}}                     & \textbf{14,771}                              & \textbf{5,276}                              & \textbf{1,055}                                     & \textbf{21,102}                              \\ \bottomrule
\end{tabular}%
}
\caption{Politifact Dataset Statistics}
\label{tab:dataset_statistics}
\end{table*}

\subsection{Politifact Dataset}
The PolitiFact dataset, introduced by \citet{misra2022politifact}, consists of 21,152 fact-checked instances (Table \ref{tab:dataset_statistics}) sourced from Politifact.com\footnotemark. Each record contains eight attributes: verdict, statement\_originator, statement, statement\_date, statement\_source, factchecker, factcheck\_date, and factcheck\_analysis\_link. The verdict classifies the truthfulness of a claim into one of six categories: \textit{true}, \textit{mostly true}, \textit{half true}, \textit{mostly false}, \textit{false}, and \textit{pants-fire}. The statements come from 13 different media categories, including speech, television, news, blog, other, social media, advertisement, campaign, meeting, radio, email, testimony, and statement. After removing instances with invalid URLs, the dataset is reduced to 21,102 instances. As false and pants on fire, both classes contain false information, so we combine these 2 classes and make a final class as false. Table \ref{table:pfo} and \ref{table:unfiltered} present the statistics for the PFO and its uncleaned subset of the PolitiFact dataset. Politi-Fact-only is the cleaned version of this uncleaned subset.

\begin{table}[h]
    \centering
    \small
    \begin{tabular}{l|cccc}
        \textbf{Label} & \textbf{Count} & \textbf{Token$_{\mu}$} & \textbf{Sent$_{\mu}$} & \textbf{BPE$_{\mu}$} \\
        \midrule
        false & 594 & 589.77 & 23.27 & 650.57 \\
        mostly-false & 600 & 808.06 & 30.30 & 890.36 \\
        half-true & 593 & 860.37 & 31.97 & 949.47 \\
        mostly-true & 598 & 765.88 & 28.84 & 847.57 \\
        true & 597 & 681.73 & 24.78 & 760.32 \\
        \midrule
        \textbf{Total} & 2982 & 741.23 & 27.83 & 819.73 \\
    \end{tabular}
    \caption{\textbf{PFO} statistics for \textit{PFO} dataset. Token$_{\mu}$, Sent$_{\mu}$, and BPE$_{\mu}$ represent the average number of standard tokens, sentences, and BPE tokens per evidence, respectively.}
    \label{table:pfo}
\end{table}

\begin{table}[h]
    \centering
    \small
    \begin{tabular}{l|cccc}
        \textbf{Label} & \textbf{Count} & \textbf{Token$_{\mu}$} & \textbf{Sent$_{\mu}$} & \textbf{BPE$_{\mu}$} \\
        \midrule
        false & 594 & 788.05 & 31.89 & 870.25 \\
        mostly-false & 600 & 1050.69 & 39.74 & 1157.32 \\
        half-true & 593 & 998.79 & 37.40 & 1103.33 \\
        mostly-true & 598 & 910.63 & 34.65 & 1008.52 \\
        true & 597 & 760.17 & 27.99 & 845.46 \\
        \midrule
        \textbf{Total} & 2981 & 901.78 & 34.34 & 997.10 \\
    \end{tabular}
    \caption{\textbf{Unfiltered} statistics for \textit{Unfiltered} dataset. Token$_{\mu}$, Sent$_{\mu}$, and BPE$_{\mu}$ represent the average number of standard tokens, sentences, and BPE tokens per evidence, respectively.}
    \label{table:unfiltered}
\end{table}

\subsection{Problem with the Politifact dataset}
The dataset contains the claim and corresponding evidence that contains the information about the claim to detect the veracity of the claim. We have some leakage in our dataset. Leakage means the evidence contains the analysis of the annotator that is published after the claim is fact-checked, which gives away the information about the label of the corresponding claim. The evidence may contain the definition of the label, some direct intuition about the label, or the label itself.

The PFO dataset is a fact-checking dataset scraped from politifact.com, focusing on the political domain. It consists of 3000 instances, each containing a political claim along with corresponding evidence. Based on the evidence, the claim’s truth value is categorized in one of the following categories: true, mostly true, half true, mostly false, false, pants on fire. We have combined pants on fire and false into one label that is false.

The dataset contains several fields, such as \textbf{Id, Label, Speaker, Claim, Evidence, Source, and Claim Date, etc.}, which are provided in the JSON file.

\textbf{Label Descriptions}
\begin{itemize}
    \item \textbf{True:} The statement is accurate and there’s nothing significant missing.
    \item \textbf{Mostly True:} The statement is accurate but needs clarification or additional information.
    \item \textbf{Half True:} The statement is partially accurate but leaves out important details or takes things out of context.
    \item \textbf{Mostly False:} The statement contains an element of truth but ignores critical facts that would give a different impression.	
    \item \textbf{False:} The statement is not accurate.
\end{itemize}	

\subsection{Instructions}
We give the following instruction to the annotator to follow while filtering the dataset.
\begin{enumerate}
    \item \textbf{Remove the " Our Ruling/Our Rating" Section:} If it exists, eliminate the section where the annotator provides their final verdict at the end of the evidence, based on the facts and analysis discussed earlier in the evidence. This section often provides explicit judgment rather than just presenting factual evidence, which can introduce biases in the automated fact-checking process.
    
    \item \textbf{Remove Sentences Containing Labels or Label Definitions:} Eliminate any sentences that directly define or provide a label (e.g., ``This claim is false'' or ``This claim is mostly true'').
    
    \item \textbf{Remove Sentences Giving Away Information About the Label:} Remove any sentences that directly reveal the label or judgment made about the claim, whether explicitly stated or implied (e.g., ``so the claim is incorrect'' or ``so the statement is partially accurate but leaves out important details'').
    
    \item \textbf{Remove Redundant Conclusions:} If the PolitiFact annotator provides a conclusion that repeats information already given in the previous sections, or that can be logically inferred from the prior content, remove it to avoid redundancy.
    
    \item \textbf{Indicate Changes in the "Leaked" Field:} Mark any evidence that requires changes by writing ``yes'' or ``no'' in the "leaked" field.
\end{enumerate}

\section{RAV pipeline analysis:}
\label{app:rav_pipeline_analysis}

\begin{figure}
    \centering
    \includegraphics[width=1\linewidth]{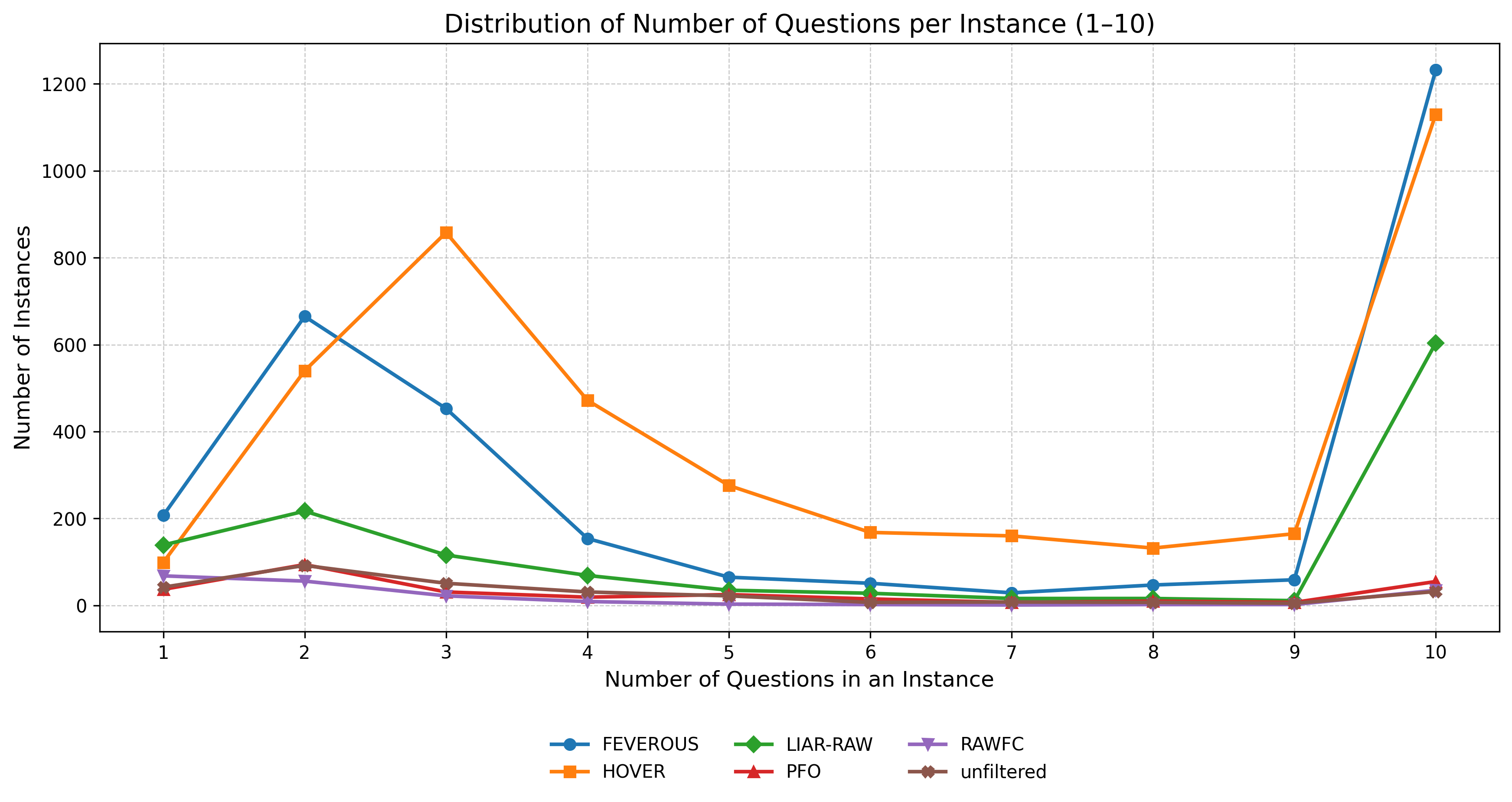}
    \caption{Distribution of number of sub-questions per Instance with reasoning Iteration \( k = 10 \). The plot shows the distribution of the number of generated questions per instance across datasets. \textsc{Feverous} and \textsc{Hover} exhibit bimodal distributions, indicating both sparse and dense questioning behaviour. In contrast, \textsc{RAWFC}, \textsc{PFO}, and unfiltered datasets are skewed toward fewer questions per instance. \textsc{LIAR-RAW} shows a right-skewed distribution peaking at 10, suggesting maximal questioning for selected samples. These patterns align with macro F1 trends, where deeper question sets correlate with improved performance.}
    \label{fig:questions_distribution}
\end{figure}

The distribution of the number of sub-questions (per instance) has been shown in Figure~\ref{fig:questions_distribution}.

\section{Human Evaluation of outputs of RAV pipeline}
\label{app:human_evaluation_of_rav}
\begin{figure*}
    \centering
    \includegraphics[width=1\linewidth]{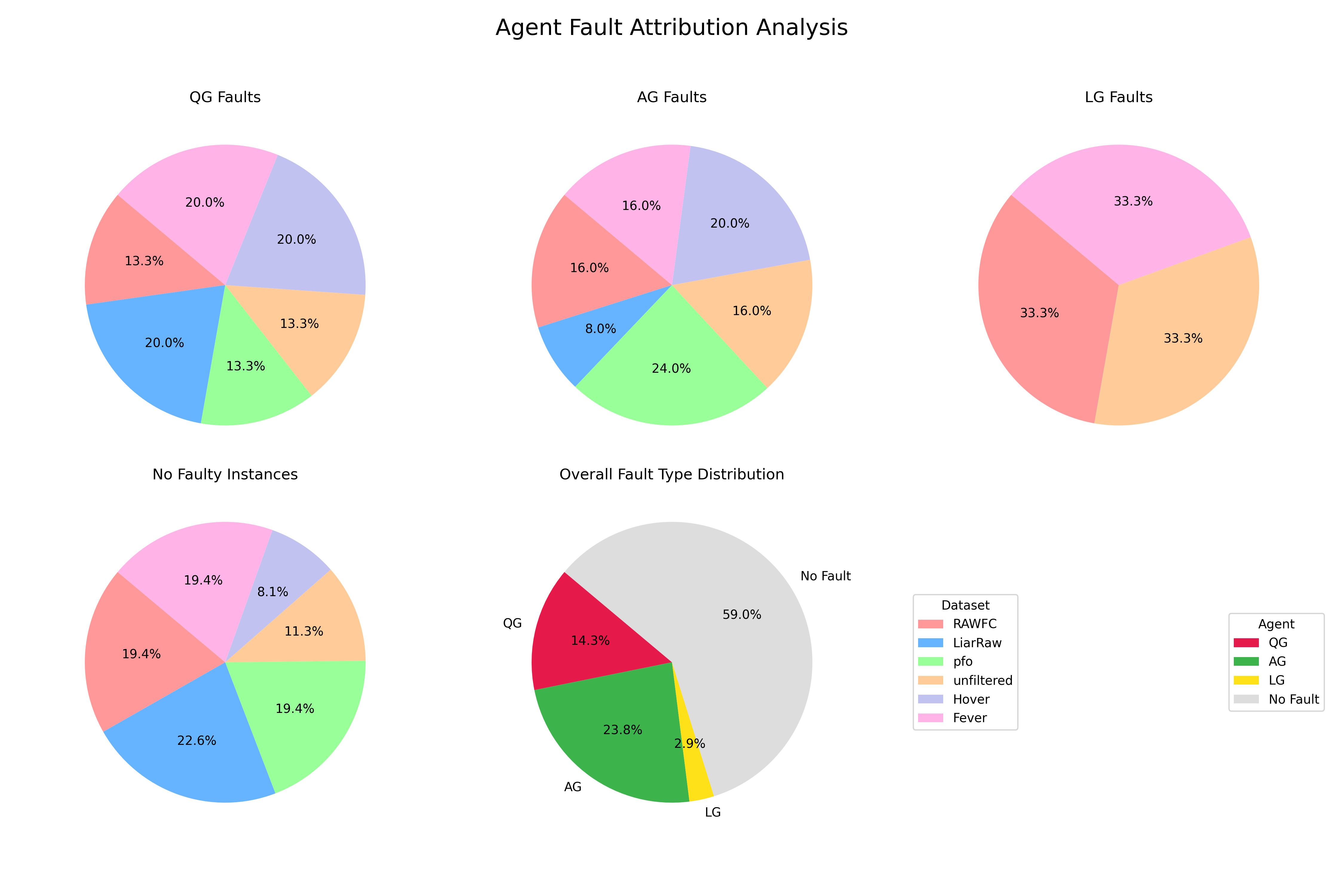}
    \caption{This figure presents the fault attribution across different datasets and agent types involved in the claim verification pipeline. The first four pie charts (top-left to bottom-left) show the proportion of instances attributed to faults in the Question Generator (QG), Answer Generator (AG), Label Generator (LG), and instances with No Fault, respectively. Only datasets with non-zero faulty instance counts are included per agent. The fifth chart (bottom-center) summarizes the overall distribution of faulty and non-faulty instances aggregated across all datasets. The sixth panel displays two legends: (left) mapping each dataset to a distinct color and (right) mapping each agent type to a unique color used in the overall distribution.}
    \label{fig:piechat}
\end{figure*}

We give 180 instances which is misclassified by the RAV pipeline for human evaluation to three annotators. This set consists of instances from all the datasets we conduct our experiments with: RAWFC, LiarRaw, PFO, PFO (Unfiltered), Hover and Feverous. Out of 180 instances, for 105 instances, all three annotators agree on the veracity label given by the RAV pipeline. We ask the evaluators to identify the fault in the RAV pipeline. The fault can be at the question generation step (QG), the answer generation step (AG) or the label generation (LG) step. We identify that on average for $\mathbf{14.29\%}$ instances the fault is at the question generation step, for $\mathbf{23.81\%}$ instances the fault is at the answer generation step, and for $\mathbf{2.86\%}$ instances the fault is at the label generation step. For the remaining $\mathbf{59.05\%}$ instances the misclassification is due to insufficient evidence. This signifies that the label generation step in RAV pipeline is able to reason well about the veracity label, given that the questions and answers are properly generated. The fault in the answer generation step can also be due to issues with evidence extraction. Although further probing is required. We show dataset-wise distribution of faulty instances in Figure~\ref{fig:piechat}.

\section{Methodology Prompts}
\label{methodology_prompt}
As discussed in Section~\ref{sec:methodology}, we provide the prompt of \(QG_{agent}\), \(AG_{agent}\), and \(LG_{agent}\). We used 8 in-context examples in the case of \(QG_{agent}\) and \(LG_{agent}\), and we do not provide any in-context examples to the answer generator. 
\begin{figure*}
    \centering
    \includegraphics[width=1\linewidth]{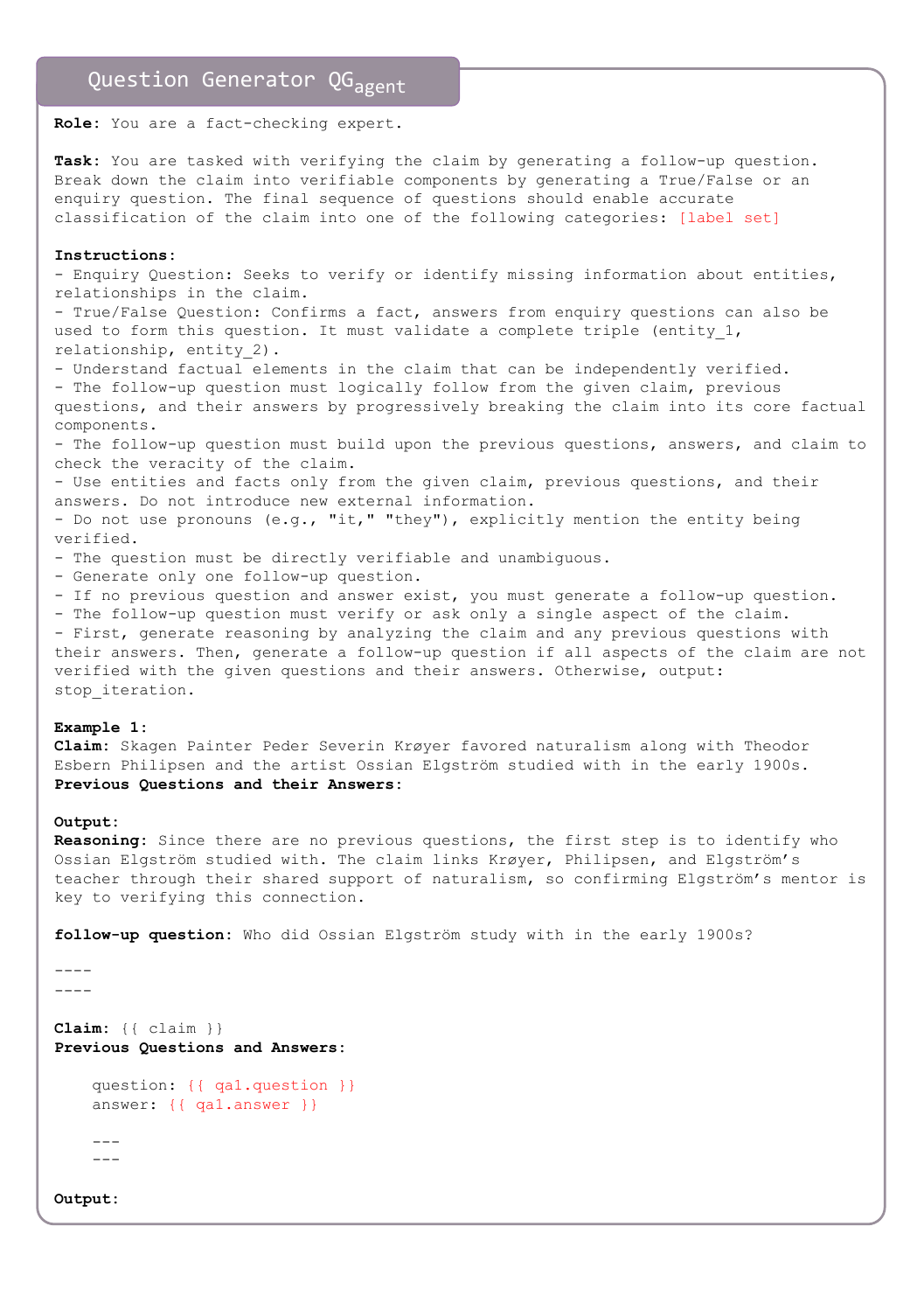}
    \caption{\(QG_{agent}\) Prompt used in our RAV pipeline, we give the clean instructions to generate the follow-up question at each iteration, stop the iteration by outputting stop\_iteration. We also provided 8 in-context examples in the prompt.}
    \label{fig:qg}
\end{figure*}

\begin{figure*}
    \centering
    \includegraphics[width=1\linewidth]{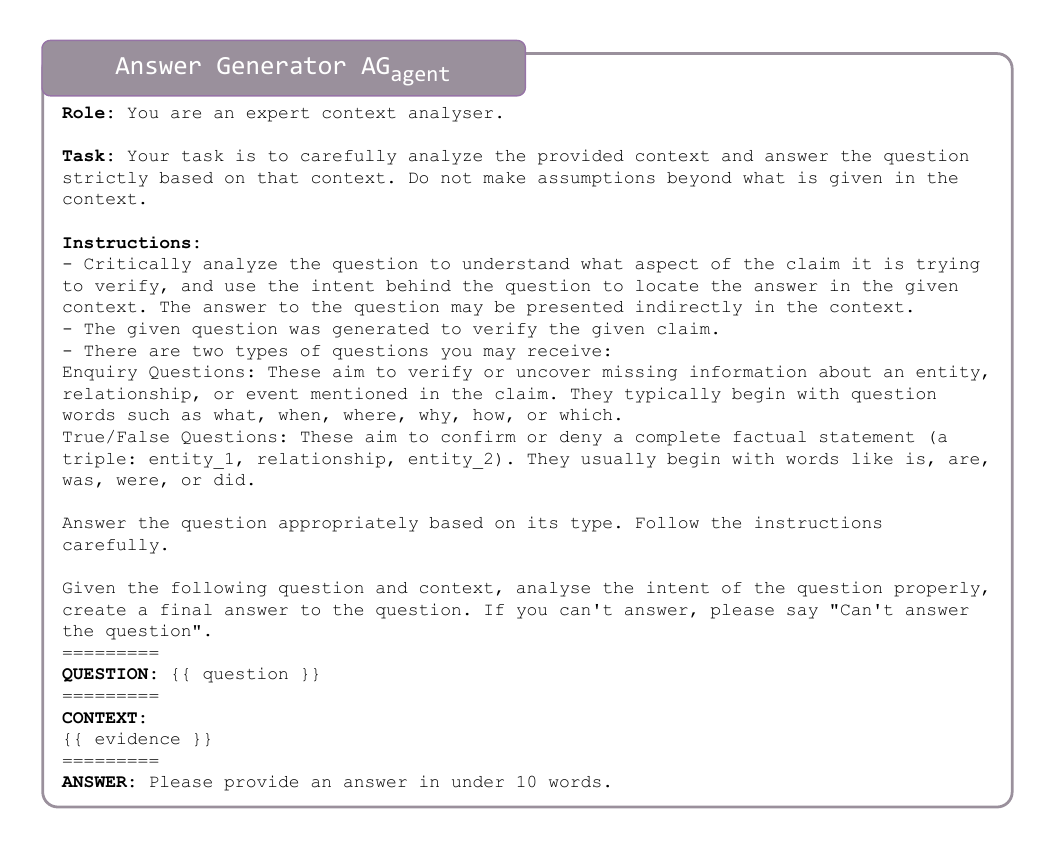}
    \caption{\(AG_{agent}\) Prompt used in our RAV pipeline, we give the clean instructions to generate the answer in 10 words, we also instruct to look for indirect answers present in the context. We also instruct \(AG_{agent}\) to completely rely on the evidence.}
    \label{fig:ag}
\end{figure*}

\begin{figure*}
    \centering
    \includegraphics[width=1\linewidth]{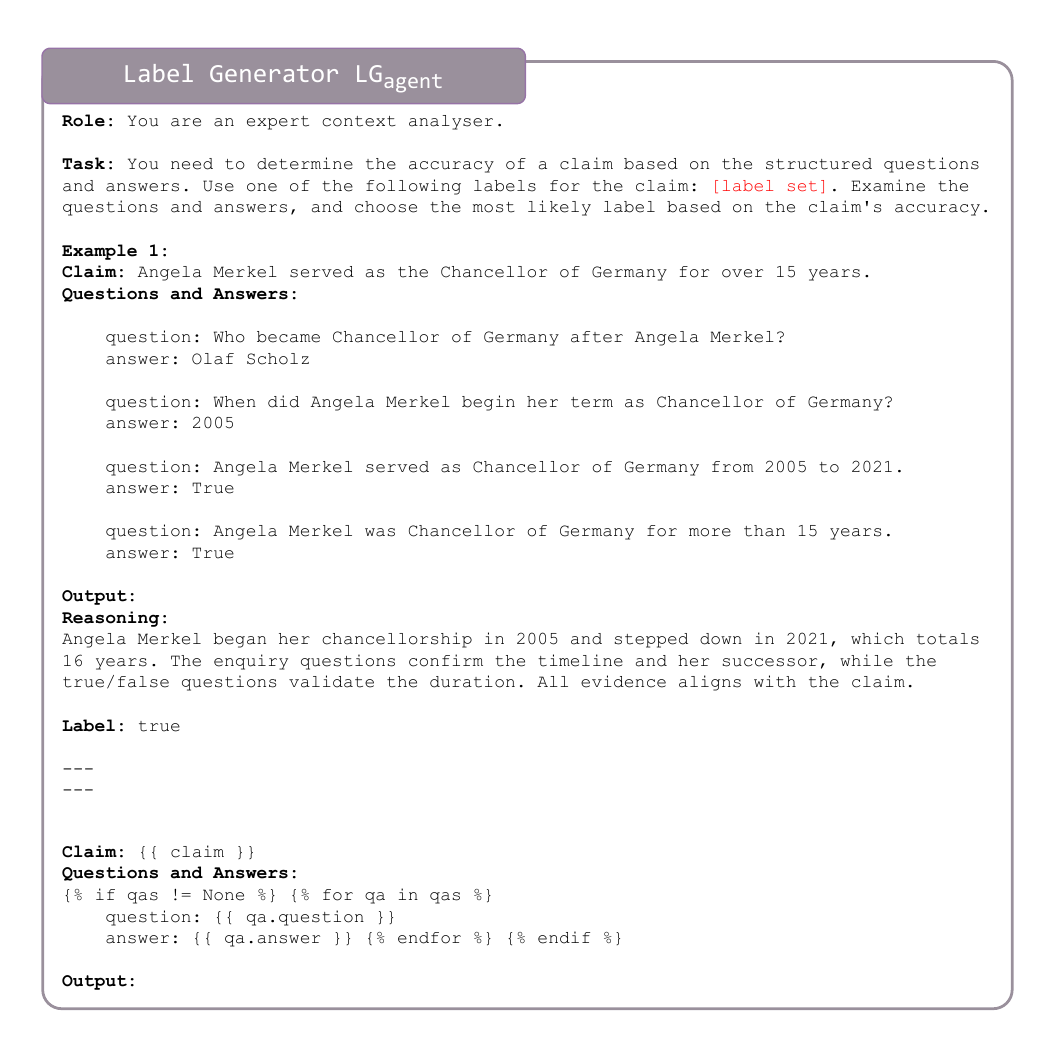}
    \caption{\(AG_{agent}\) Prompt used in our RAV pipeline, we give a short description of the task and provide 8 in-context examples in the prompt, we instruct \(AG_{agent}\) to first generate a reasoning to connect the claim and generated questions and answers and then predict the label}
    \label{fig:lg}
\end{figure*}

\section{Examples from PFO dataset}
\label{appex:pfo_examples}
\begin{figure*}[t]
    \centering
    \includegraphics[width=1\linewidth]{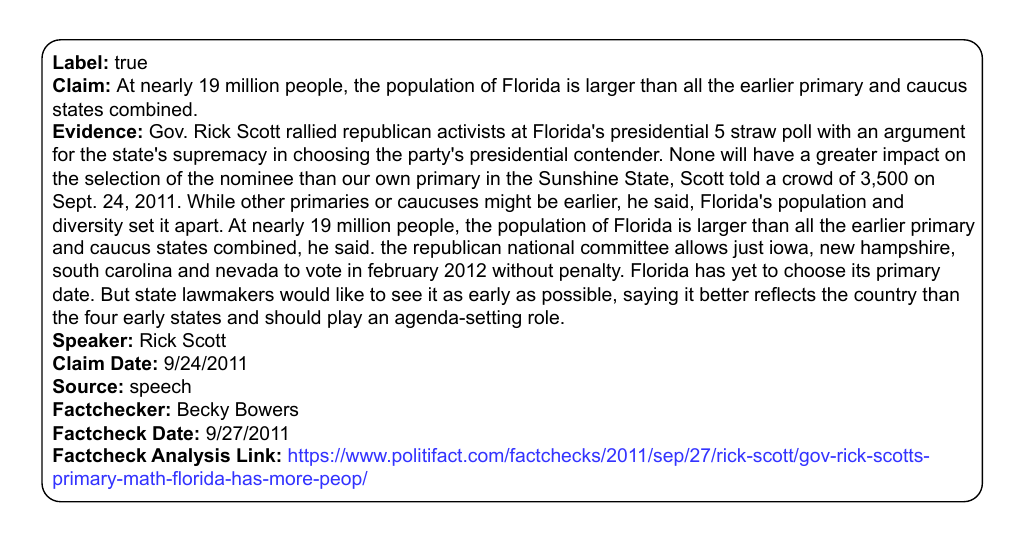}
    \caption{A true instance from the PFO dataset.}
    \label{fig:true}
\end{figure*}

\begin{figure*}[t]
    \centering
    \includegraphics[width=1\linewidth]{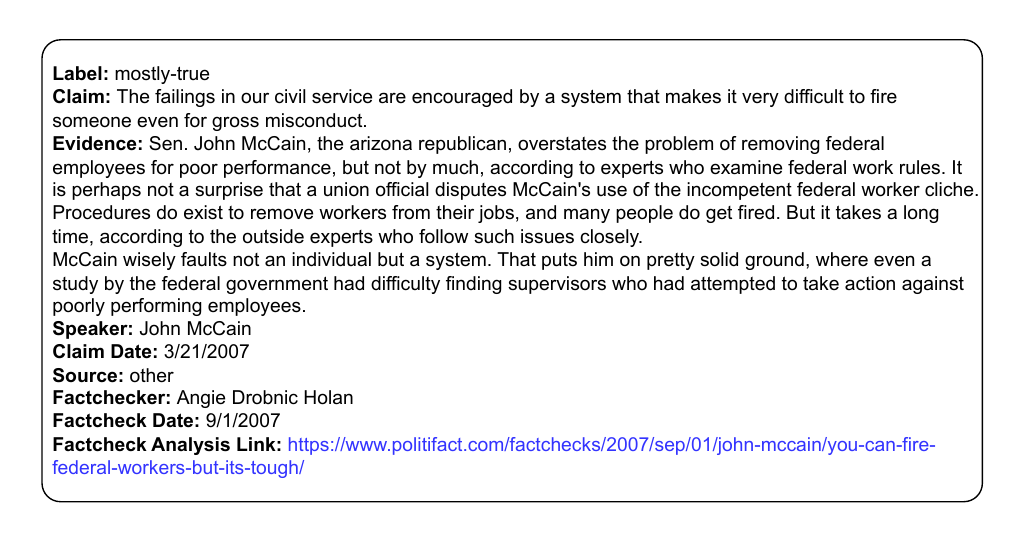}
    \caption{A mostly true instance from the PFO dataset.}
    \label{fig:mostly_true}
\end{figure*}

\begin{figure*}[t]
    \centering
    \includegraphics[width=1\linewidth]{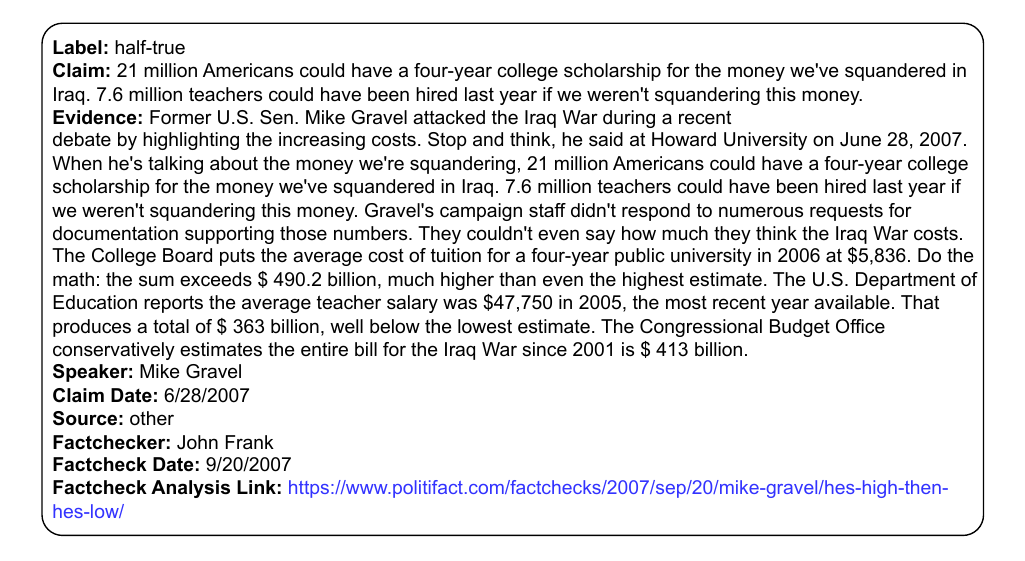}
    \caption{A half true instance from the PFO dataset.}
    \label{fig:half_true}
\end{figure*}

\begin{figure*}[t]
    \centering
    \includegraphics[width=1\linewidth]{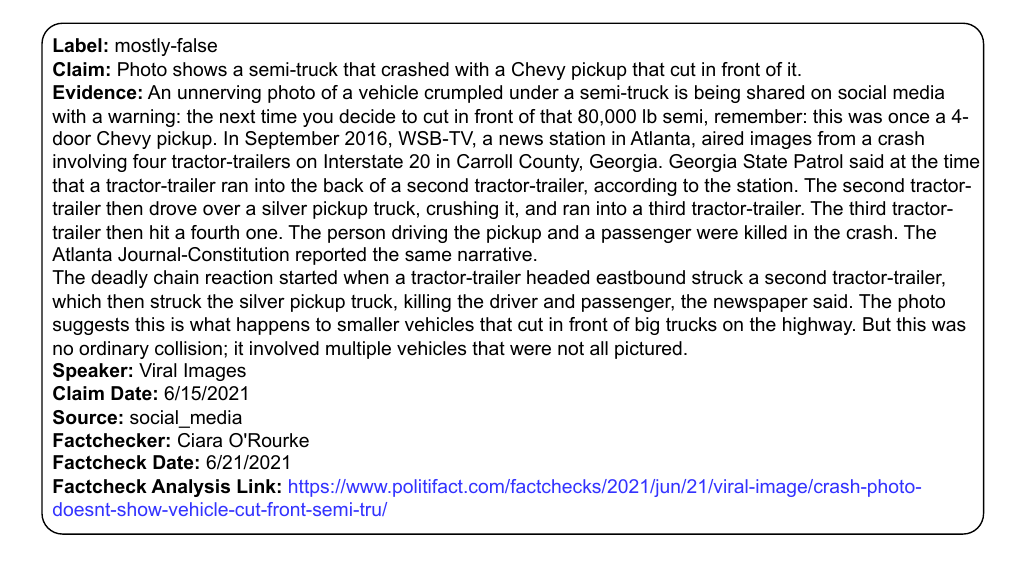}
    \caption{A mostly false instance from the PFO dataset.}
    \label{fig:mostly_false}
\end{figure*}

\begin{figure*}[t]
    \centering
    \includegraphics[width=1\linewidth]{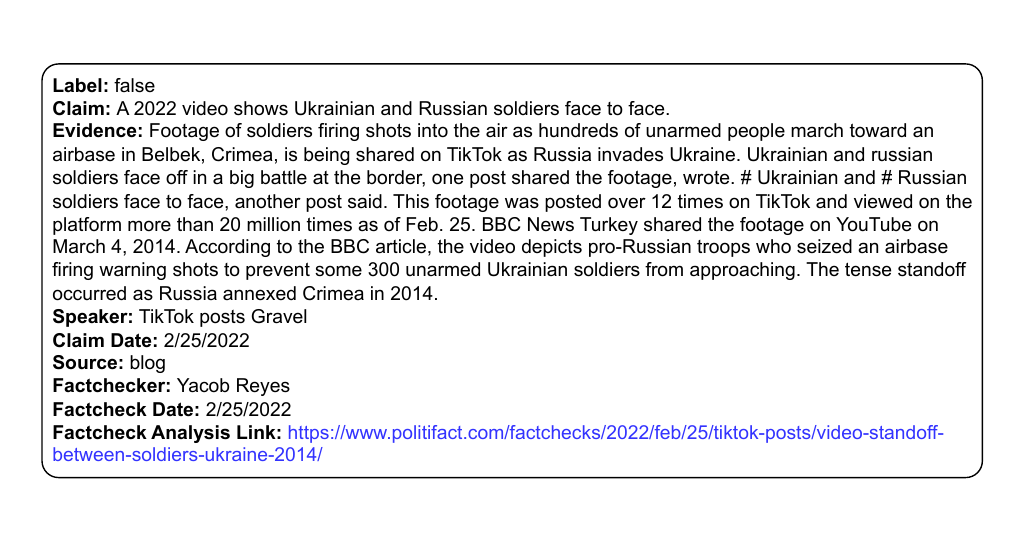}
    \caption{A false instance from the PFO dataset.}
    \label{fig:false}
\end{figure*}

As discussed in Section~\ref{sec:dataset}, we include one example from each class in the PFO dataset in Figures~\ref{fig:true} to~\ref{fig:false} to illustrate the content of our proposed dataset.

\end{document}